\documentclass[]{article} 
\usepackage[]{aaai23}  
\usepackage{times}  
\usepackage{helvet}  
\usepackage{courier}  
\usepackage[hyphens]{url}  
\usepackage{graphicx} 
\usepackage{natbib}  
\usepackage{caption} 
\usepackage[algo2e]{algorithm2e}
\usepackage{subcaption}
\usepackage{algorithm}
\usepackage{color}

\usepackage{newfloat}
\usepackage{listings}
\lstdefinelanguage{PDDL}
{
  sensitive=false,    
  morecomment=[l]{;}, 
  alsoletter={:,-},   
  morekeywords={
    define,domain,problem,not,and,or,when,forall,exists,either,
    :domain,:requirements,:types,:objects,:constants,
    :predicates,:action,:parameters,:precondition,:effect,
    :fluents,:primary-effect,:side-effect,:init,:goal,
    :strips,:adl,:equality,:typing,:conditional-effects,
    :negative-preconditions,:disjunctive-preconditions,
    :existential-preconditions,:universal-preconditions,:quantified-preconditions,
    :functions,assign,increase,decrease,scale-up,scale-down,
    :metric,minimize,maximize,
    :durative-actions,:duration-inequalities,:continuous-effects,
    :durative-action,:duration,:condition
  }
}
\DeclareCaptionStyle{ruled}{labelfont=normalfont,labelsep=colon,strut=off} 
\floatstyle{ruled}
\newfloat{listing}{tb}{lst}{}
\floatname{listing}{Listing}
\usepackage{multirow}

\title{On Solving the Rubik's Cube with Domain-Independent Planners \\ Using Standard Representations}
\author {
    Bharath Muppasani,
    Vishal Pallagani,
    Biplav Srivastava,
    Forest Agostinelli
}
\affiliations {
    AI Institute, University of South Carolina, Columbia, South Carolina, USA\\
    \{bharath@email., vishalp@email., 
    biplav.s@, foresta@cse.\}sc.edu
}

\usepackage{bibentry}

\begin{document}
%


\maketitle
\begin{abstract}

\begin{quote}

Rubik's Cube (RC) is a well-known and computationally challenging puzzle that has motivated AI researchers to explore  efficient alternative representations and problem-solving methods. The ideal situation for planning here is that a problem be solved optimally and efficiently represented in a standard notation using a general-purpose solver and heuristics. The fastest solver today for RC is DeepCubeA with a custom representation, and another approach is with Scorpion planner with State-Action-Space+ (SAS+) representation. In this paper, we present the first RC representation in the popular PDDL language so that the domain becomes more accessible to PDDL planners, competitions, and knowledge engineering tools, and is more human-readable. We then bridge across existing approaches  and compare performance.
We find that in one comparable experiment, DeepCubeA\footnote{DeepCubeA trained with 12 RC actions} solves all problems with varying complexities, albeit only 78.5\% are optimal plans. For the same problem set, Scorpion with SAS+ representation and pattern database heuristics solves 61.50\% problems optimally, while FastDownward with PDDL representation and FF heuristic solves 56.50\% problems, out of which 79.64\% of the plans generated were optimal. Our study provides valuable insights into the trade-offs between representational choice and plan optimality that can help researchers design future strategies for challenging domains combining general-purpose solving methods (planning, reinforcement learning), heuristics, and representations (standard or custom).

\end{quote}
\end{abstract}

\section{Introduction}

The Rubik's Cube is a 3D puzzle game that has been widely popular since its invention in 1974. It has been a subject of interest for researchers in Artificial Intelligence (AI) due to its computational complexity and potential for developing efficient problem-solving algorithms. RC has motivated researchers to explore alternative representations that simplify the problem while preserving its complexity. Efficient algorithms have been developed to solve RC in the least number of moves, and they have been used in various applications, including robot manipulation, game theory, and machine learning. Therefore, in this paper, we aim to explore the different representations and algorithms to solve RC and evaluate their performance and effectiveness in solving this challenging puzzle.



Various solution approaches  have been proposed RC including Reinforcement Learning (RL) and search.
For instance, DeepCubeA \cite{agostinelli2019solving} uses RL to learn policies for solving RC, where the cube state is represented by an array of numerical features. Although DeepCubeA is a domain-independent puzzle solver, it employs a custom representation for RC. On the other hand, \citet{buchner2022comparison} utilized SAS+ representation to model the RC problem in a finite domain representation, which enables standard general-purpose solvers like Scorpion to be used on the RC problem. Despite the success of these approaches, no prior work has explored the use of Planning Domain Definition Language (PDDL) to encode a 3x3x3 RC problem. While a previous study\footnote{https://wu-kan.cn/2019/11/21/Planning-and-Uncertainty/} has encoded a 2x2x2 RC problem using PDDL and solved it with a Fast-Forward planner, there exists no PDDL encoding for a 3x3x3 RC problem.

\begin{table}[!t]
\centering
\begin{tabular}{|l|l|}
\hline
\textbf{Notations} & \textbf{Description} \\ \hline
RC & Rubik's Cube \\ \hline 
 \hline
PDDL &  Planning Domain Description Language \\ 
 & \cite{pddl2.1}\\ \hline
SAS+ &  \begin{tabular}[c]{@{}l@{}}State-Action-Space+ \\ \cite{fikes1971strips} \end{tabular}\\ \hline
Custom & \begin{tabular}[c]{@{}l@{}}RC representation in DeepCubeA \\ \cite{agostinelli2019solving} \end{tabular}\\ \hline \hline
Blind & FastDownward with Blind\textsuperscript{3} \\ \hline
GC & FastDownward with Goal count\textsuperscript{3} \\ \hline
CG & FastDownward with Causal Graph\textsuperscript{3} \\ \hline
CEA & \begin{tabular}[c]{@{}l@{}}FastDownward with \\ Context-enhanced Additive\textsuperscript{3}\end{tabular} \\ \hline
LM-Cost & \begin{tabular}[c]{@{}l@{}}FastDownward with \\ LM-Cost Partitioning\textsuperscript{3}\end{tabular} \\ \hline
FF & FastDownward with FF\textsuperscript{3}\\ \hline \hline
M\&S & Scorpion with Merge \& Shrink\textsuperscript{4}\\ \hline
PDB-Man & Scorpion with Max Manual PDB\textsuperscript{4}\\ \hline
PDB-Sys & Scorpion with Max Systematic PDB\textsuperscript{4} \\ \hline \hline
d1 & \begin{tabular}[c]{@{}l@{}}Dataset of 200 problems generated \\ considering 12 RC actions\end{tabular} \\ \hline
d2 & \begin{tabular}[c]{@{}l@{}}Dataset of 200 problems generated \\ considering 18 RC actions\end{tabular} \\ \hline \hline
m1 & PDDL model with 12 RC actions \\ \hline
m2 & PDDL model with 18 RC actions \\ \hline
\end{tabular}
\caption{Notations or abbreviations and their descriptions.}
\label{tab:notations}
\end{table}

In this paper, we introduce a novel approach for representing RC in PDDL. We encode the initial state and goal state using a set of predicates, each of which specifies the color of a sticker on a particular cube piece or edge piece. We then define the actions that can be taken to manipulate the cube pieces and edges. Our PDDL representation enables us to model RC as a classical planning problem, which can be solved using off-the-shelf planning tools. To the best of our knowledge, this is the first attempt to represent RC formally using PDDL. We also evaluate the effectiveness of our approach by comparing it with other state-of-the-art representations in terms of the efficiency and effectiveness of problem-solving.
Our major contributions are:
\begin{itemize}

    \item We develop the first PDDL formulation for the 3x3x3 Rubik's Cube, which is a novel and significant contribution to the existing literature. This PDDL formulation will enable the use of standard PDDL planners for solving Rubik's Cube problems, which was not previously possible.
     \item We bridge across hither-to incomparable RC solving approaches, compare their performance  and draw insights from results to facilitate new research.
    \item We perform a comparative analysis of two formal languages, SAS+ and PDDL, and custom one in DeepCubeA, a RL approach for solving RC on a set of common benchmark RC problems. This comparative analysis is important as it provides insights into the strengths and weaknesses of these different approaches, and helps to identify which method may be most appropriate for a given problem setting.
\end{itemize}

The paper is organized as follows: we begin with giving an overview of Rubik's Cube solving ecosystem, including the RC problem, domain-independent planners and heuristics, and learning-based RC solvers. Then, we present a comparison of three different representations for RC: DeepCubeA, SAS+, and PDDL. Next, we outline the experiments conducted, including the heuristics considered and the experimental setup, followed by  results. We compare RC solvers and heuristics for the number of problems solved and plan optimality. Finally, the paper concludes with a discussion of the findings and their implications for future research in solving larger RC problems.

\footnotetext[3]{https://www.fast-downward.org/Doc/Evaluator}
\footnotetext[4]{https://jendrikseipp.github.io/scorpion/Evaluator/}
\section{The RC Solving Ecosystem}

\begin{figure}[!t]
	\centering
	\includegraphics[scale=0.1, width=0.40\textwidth]{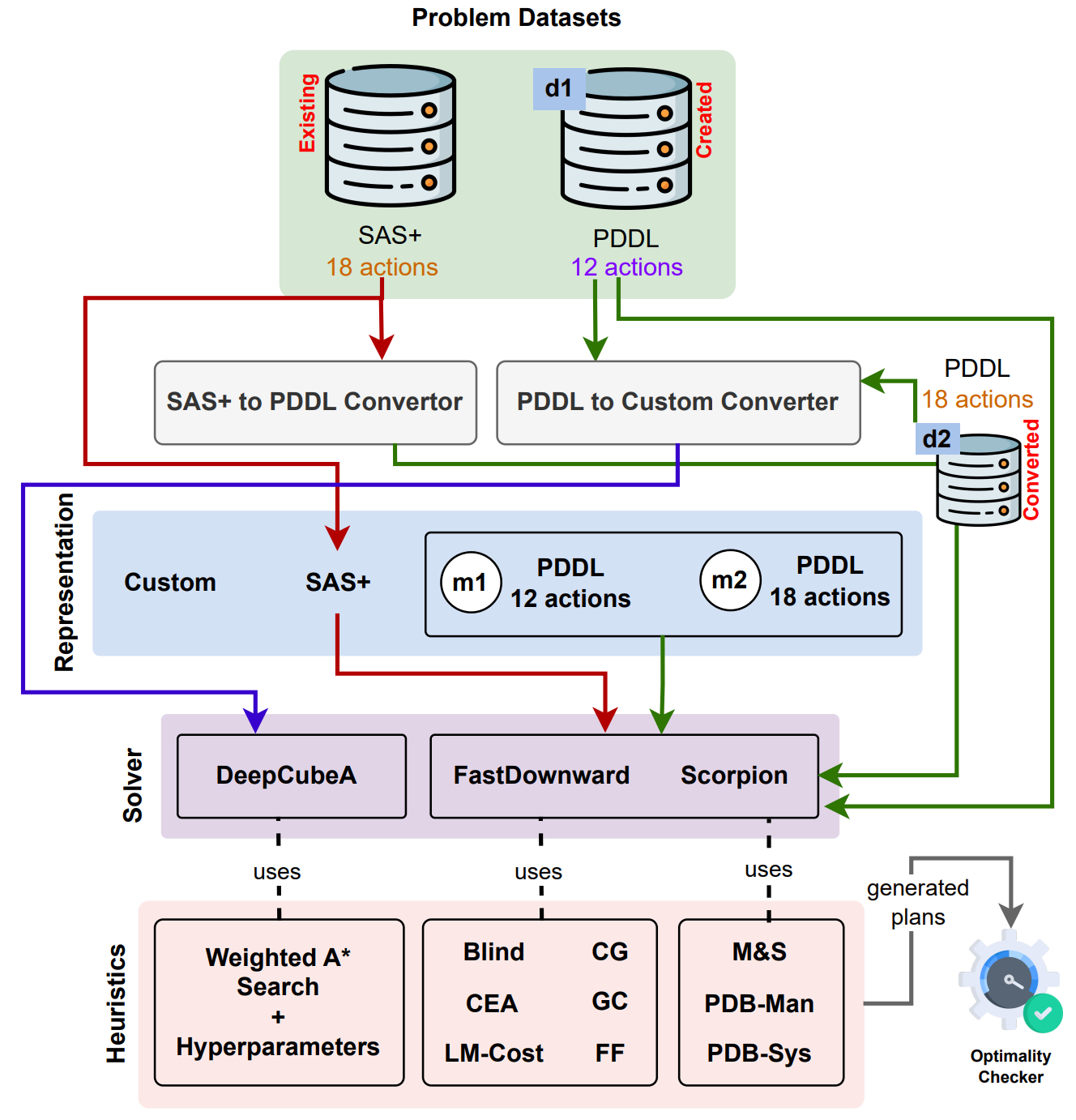}
	\caption{Setup for comparing RC representation, solvers and heuristics. Converters crucially bridge representations.} 
	\label{FIG: ecosystem}
\end{figure}

In this section, we describe the RC problem, planners, and heuristics that are used for our study. Table~\ref{tab:notations} summarizes the notations used and Figure \ref{FIG: ecosystem} shows the entire functionality of our designed ecosystem.

\subsection{RC Problem}
The Rubik's Cube is a 3-D combination puzzle with colored faces made up of 26 smaller colored pieces linked to a central spindle, with the goal of rotating the blocks until each face of the cube is a single color. To solve the puzzle, one can perform certain actions that correspond to the different faces of the cube. The major actions of a Rubik's cube are Up(U), Down(D), Right(R), Left(L), Front(F), and Back(B), which define a rotation of 90 degrees in a clockwise direction of the respective face per action. The inverse of these actions corresponds to a 90-degree rotation in the anti-clockwise direction (suffix 'rev'). The cube is initially rotated by a random sequence of rotations in the puzzle's initial configuration. The goal is to find a series of rotations that results in the solved state, which has all faces displaying the same color. One can solve the RC from a scrambled state to the original solved configuration by performing a set of the above-mentioned actions. 
\subsection{Domain-Independent Planners and Heuristics}

\subsubsection{Classical Planning Formalism} 
Consider $F$ to be a set of propositional variables or \textit{fluents}. A \textit{state} $s \subseteq F$ is a subset of fluents that are true, while all fluents in $F \setminus s$ are implicitly assumed to be false. A subset of fluents $F^\prime \subseteq F$ holds in a state $s$ if and only if $F^\prime \subseteq s$. A classical planning instance is a tuple $P = \langle F, A, I, G \rangle$, where $F$ is a set of fluents, $A$ is a set of actions, $I \subseteq F$ an initial state, and $G \subseteq F$ a goal condition. Each action $a \in A$ has precondition pre($a$) $\subseteq F$, add effect add($a$) $\subseteq F$, and delete effect del($a$) $\subseteq F$, each a subset of fluents. Action $a$ is applicable in state $s \subseteq F$ if and only if pre($a$) holds in $s$, and applying $a$ in $s$ results in a new state $s \oplus a = (s \setminus$ del($a$)) $\cup$ add($a$). A \textit{plan} for $P$ is a sequence of actions $\prod = \langle a_{1},...,a_{n} \rangle$ such that $a_{1}$ is applicable in $I$ and, for each $2 \le i \le n, a_{i}$ is applicable in $I \oplus a_{1} \oplus ... \oplus a_{i-1}$. The plan $\prod$ solves $P$ if $G$ holds after applying $a_{1},...,a_{n}$, i.e. $G \subseteq I \oplus a_{1} \oplus ... \oplus a_{n}$.

\subsubsection{Abstractions}
Let $\mathcal{T} = \langle S, \mathcal{L}, T, s_{I}, S_{*} \rangle$ be a transition system. An \textit{abstraction} $\alpha : S \rightarrow S^{\alpha}$ maps the states of $\mathcal{T}$ to a set of \textit{abstract states} $S^{\alpha}$. The induced transition system is $\mathcal{T}^{\alpha} = \langle S_{\alpha}, \mathcal{L}, T^{\alpha}, \alpha(s_{I}), \{\alpha(s) | s \in S_{*}\} \rangle$ where $T^{\alpha} = \{\langle \alpha(s),o,\alpha(s^{\prime}) \rangle | \langle s, o, s^{\prime} \rangle \in T\}$. By construction, every path in $\mathcal{T}$ is a path in $\mathcal{T^{\alpha}}$. Consequently, the length of the shortest path between state $\alpha(s)$ and $\alpha(s^{\prime})$ in $\mathcal{T}^{\alpha}$ is a lower bound on the length of the shortest path between state $s$ and $s^{\prime}$ in $\mathcal{T}$. Thus, the abstract goal distance for a given state is an admissible estimate of the true goal distance \citep{buchner2022comparison}. In the later section of the paper, we mention the abstraction heuristics used for our work.
\subsubsection{PDDL}

The field of planning has seen many representations. For example, in classical planning, there was STRIPS~\cite{strips}, Action Description Language (ADL)~\cite{adl} and SAS+~\cite{comparison-pl-backstorm} before Planning Domain Description Language (PDDL)~\cite{pddl,pddl2.1} standardized the notations. Nowadays, planners routinely use PDDL for problem specification even if they may convert to other representations later for solving  efficiency \cite{pddl-conversion-finite}.  PDDL envisages two files, a domain description file which specifies information independent of a problem like predicates and actions, and a problem description file which specifies the initial and goal states. A problem is characterized by an initial state, together with a goal state that the agent wants to transition to, both states specified as configurations of objects. A planner takes as input the domain and problem file to generate a plan, which can be verified using a plan validator, VAL \cite{plan-verif-val}.





\subsection{Learning-based RC Solver}

There exist specialized solvers for solving the Rubik's Cube, which can be classified as either domain-dependent or domain-independent. DeepCubeA is an example of a domain-independent solver that employs custom representation encoding for RC, as proposed in \cite{mcaleer2018solving,deepcube}. The solver employs a weighted A* search algorithm and learns a domain-dependent heuristic via deep-learning, resulting in state-of-the-art performance. However, interpreting the solutions provided by DeepCubeA remains a challenge \cite{deepcube-iclr}, as does comparing its performance with that of other solvers {\em on identical problem instances}.

\section{Comparision of RC representations}

In this section, we describe and provide a comparative analysis of different RC representations comprising of RL and Planning formal languages.
\subsection{DeepCubeA}
The DeepCubeA algorithm adopts a unidimensional array as a representation of the Rubik's Cube (RC) state. Specifically, this array encompasses 54 elements, each of which corresponds to a unique sticker color present on a cube piece of the RC. While this array-based modeling offers computational advantages, it is limited by its inability to fully encapsulate the spatial orientation of Rubik's Cube. Furthermore, the usage of a hard-coded representation and implicit assumptions concerning the position of cubelets poses a challenge to novice users seeking to comprehend the array-based representation.
\subsection{SAS+}
In \citealt{buchner2022comparison}, Rubik's Cube is modeled with 18 actions in SAS+ representation as a factored effect task, with each face labeled as F, B, L, R, U, or D. The orientation of each cube piece is represented as a triple of values, and for corner cube pieces, the orientation is a permutation of \{1, 2, 3\}, while for edge cube pieces, it is a permutation of \{1, 2, \#\} (where \# represents a blank symbol). The rotation of the cube in 3D space is captured as the permutation of the respective triple for each cube piece. The SAS+ model has 20 variables, 480 fact pairs and bytes required for representing each state is 16 bytes.

\begin{figure}[!b]
 \centering
 \includegraphics[scale=0.9]{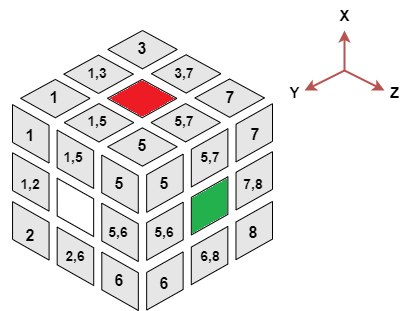}
 \caption{Rubik's cube description to define the domain encoding.}
 \label{FIG: rubiks-cube}
\end{figure}


\begin{lstlisting}[
  float=t,
  caption={Action L of Rubik's Cube modeled in PDDL},
  label={lst:pddl},
  language=PDDL, numbers=none,
  frame=lines,
  xleftmargin=\parindent, xrightmargin=\parindent
  ]
(:action L
:effect (and
;for corner cubelets
(forall(?x ?y ?z)(when (cube1 ?x ?y ?z) 
  (and (cube2 ?y ?x ?z))))
(forall(?x ?y ?z)(when (cube3 ?x ?y ?z) 
  (and (cube1 ?y ?x ?z))))
(forall(?x ?y ?z)(when (cube4 ?x ?y ?z) 
  (and (cube3 ?y ?x ?z))))
(forall(?x ?y ?z)(when (cube2 ?x ?y ?z) 
  (and (cube4 ?y ?x ?z))))
;for edge cubelets
(forall(?x ?z)(when (edge13 ?x ?z) 
  (and (edge12 ?x ?z))))
(forall(?y ?z)(when (edge34 ?y ?z) 
  (and (edge13 ?y ?z))))
(forall(?x ?z)(when (edge24 ?x ?z) 
  (and (edge34 ?x ?z))))
(forall(?y ?z)(when (edge12 ?y ?z) 
  (and (edge24 ?y ?z))))))
\end{lstlisting}
\subsection{PDDL}

In the PDDL domain, the Rubik's cube problem environment has been defined by assuming the cube pieces are in a fixed position and are named accordingly, as defined in Figure \ref{FIG: rubiks-cube}. These fixed cube pieces are modeled as predicates in the RC domain and the colors they possess in the three-dimensional space as parameters of these predicates. With the help of conditional effects, each action in the RC environment is defined as the change of colors on these fixed cube pieces. The 3D axis of the cube is considered as three separate parameters {\em X}, {\em Y}, and {\em Z} that specify the position of the colors on the cube's pieces. One of these axes can be connected to each face of the cube. According to the representation shown in Figure \ref{FIG: rubiks-cube}, the respective faces on each axis are: $F_{X} = \langle U,D \rangle$; $F_{Y} = \langle R,L \rangle$; $F_{Z} = \langle F,B \rangle$. These different faces of the cube can be identified by the color of the middle cube piece. We considered White, Red, and Green colors as the colors on the front(F), up(U) and right(R) faces respectively (similarly, the counter colors on the counter faces).

The following conventions regarding the RC cube pieces are considered to model the RC domain actions in the PDDL:
\begin{enumerate}
  \item The corner cube pieces of the RC are modeled as a three-color cubelet and are specified as a predicate with three parameters: {\em x}, {\em y}, and {\em z}, which indicate the piece's colors on three separate axes. There are 8 corner pieces in RC.
  \item The edge cube pieces, which are in between corner cube pieces, are modeled as two-color cubelet and is specified as a predicate with two parameters denoting the piece's colors on the two axes. There are 12 edge pieces in RC.
  \item We do not consider the rotations performed on the middle layer, as this can be resolved into rotation of right and left faces in the opposite direction. As a result, the middle cube piece of a face is unaltered.
\end{enumerate}

The predicate names define the fixed position of the cubelets that are defined with respect to the different faces of the cube. The representation considered for the cube positions is shown in Figure \ref{FIG: rubiks-cube}. One of the actions, action `L', of RC designed in PDDL from the description provided is shown in Listing \ref{lst:pddl}. In this, we refer to corner cube pieces as \textit{cubeP} and edge cube pieces as \textit{edgePQ} where \textit{P} and \textit{Q} are the numbers for the cube pieces as stated in Figure \ref{FIG: rubiks-cube}.
When the move L is applied to the RC, for example, the left face is rotated clockwise. This may be regarded as a 90-degree clockwise translation of colors from the left-face corner and edge cube pieces. Considering the RC representation shown in Figure \ref{FIG: rubiks-cube}, the colors on the pieces: cube1, cube2, cube4, and cube3, are circularly shifted towards the right. The same applies to the edge pieces. As the left face falls in the Z-plane, only the  X-axis and Y-axis colors on the cube pieces are affected. 

During execution of a problem, FastDownward first translates the domain into a SAS model. The resulting SAS version of the RC PDDL domain model has 480 variables and 960 fact pairs, with each state requiring 60 bytes for representation. 

\section{Experiments}

\begin{table*}[!t]
\centering
\begin{tabular}{|l|cc|cc|c|}
    \hline
    \multirow{2}{*}{\textbf{Planner with Heuristic}}  & \multicolumn{2}{c|}{\textbf{d1}} & \multicolumn{3}{c|}{\textbf{d2}} \\ \cline{2-6} 
     & \multicolumn{1}{c|}{\textbf{m1}} & \multicolumn{1}{c|}{\textbf{m2}} & \multicolumn{1}{c|}{\textbf{m1}} & \multicolumn{1}{c|}{\textbf{m2}} & \multicolumn{1}{c|}{\textbf{SAS+\textsuperscript{*}}} \\ \hline
    
    FastDownward with Blind & \multicolumn{1}{c|}{78 \textit{(100\%)}} & 67 \textit{(100\%)}& \multicolumn{1}{c|}{56 \textit{(100\%)}} & 65 \textit{(100\%)} & 66 \textit{(100\%)}\\ 
    
    
    FastDownward with Causal Graph & \multicolumn{1}{c|}{96 \textit{(66.67\%)}} & 76 \textit{(81.58\%)} & \multicolumn{1}{c|}{72 \textit{(68\%)}} & 75 \textit{(80\%)} & 77 \textit{(75.32\%)} \\ 
    
    FastDownward with Context-enhanced Additive & \multicolumn{1}{c|}{99 \textit{(61.62\%)}} & 71 \textit{(85.92\%)} & \multicolumn{1}{c|}{68 \textit{(66.18\%)}} & 75 \textit{(78.67\%)} & 77 \textit{(75.32\%)} \\ 
    
    FastDownward with Goal count & \multicolumn{1}{c|}{103 \textit{(99.03\%)}} & 88 \textit{(97.73\%)} & \multicolumn{1}{c|}{75 \textit{(94.67\%)}} & 85 \textit{(89.41\%)} & 87 \textit{(89.66\%)} \\ 
    
    FastDownward with LM-Cost Partitioning & \multicolumn{1}{c|}{103 \textit{(100\%)}} & 97 \textit{(100\%)} & \multicolumn{1}{c|}{75 \textit{(100\%)}} & 86 \textit{(100\%)} & 87 \textit{(100\%)} \\ 
     
    \textbf{FastDownward with FF }& \multicolumn{1}{c|}{\textbf{137} \textit{(99.27\%)}} & \textbf{135} \textit{(88.89\%)} & \multicolumn{1}{c|}{\textbf{104} \textit{(98\%)}} & \textbf{113} \textit{(79.64\%)} & \textbf{123} \textit{(79.67\%)} \\ 
    
    \hline
    \hline
    Scorpion with Merge \& Shrink & \multicolumn{1}{c|}{114 \textit{(100\%)}} & 105 \textit{(100\%)} & \multicolumn{1}{c|}{82 \textit{(100\%)}} & 95 \textit{(100\%)} & 90 \textit{(100\%)}\\
    
    Scorpion with Max Manual PDB & \multicolumn{1}{c|}{95 \textit{(100\%)}} & 83 \textit{(100\%)} & \multicolumn{1}{c|}{64 \textit{(100\%)}} & 78 \textit{(100\%)} & 123 \textit{(100\%)} \\
    
    Scorpion with Max Systematic PDB & \multicolumn{1}{c|}{88 \textit{(100\%)}} & 78 \textit{(100\%)} & \multicolumn{1}{c|}{63 \textit{(100\%)}} & 73 \textit{(100\%)} & 120 \textit{(100\%)} \\ 
    
    \hline
    \hline
    DeepCubeA (12 action model) & \multicolumn{2}{c|}{200 \textit{(94.5\%)}} & \multicolumn{3}{c|}{200 \textit{(78.5\%)}}  \\
    \hline
\end{tabular}
\caption{Comparison of planner configurations based on the total number of solved problems and the percentage of optimal plans for different Rubik's Cube models. (\textsuperscript{*}SAS+ dataset presented by \citet{buchner2022comparison})}
\label{tab:exp-results}
\end{table*}

In the following section, we will discuss the heuristics considered in our evaluation and the experimental setup, which includes the datasets, problem representations, and details about the planner.


\subsection{Heuristics Considered}


\subsubsection{Blind Heuristic} refers to a decision-making strategy that does not incorporate any specific information regarding the problem domain. It relies solely on the present state of the problem and employs a trial-and-error method to find a solution.
        
\subsubsection{Causal Graph Heuristic} utilizes the causal relationships between state variables in a planning task. This graph captures dependencies among variables, with arcs indicating the influence between them. By leveraging the graph's structure, the heuristic offers insights into variable interactions, facilitating more efficient planning in domains where causal understanding is pivotal. \cite{helmert2004planning}

\subsubsection{Context-enhanced Additive Heuristic} refines the additive heuristic by using contextual information, offering a more precise estimate. While the additive heuristic determines costs based on individual goals, the context-enhanced version evaluates conditions within distinct contexts, differing from the seed state's heuristic value. Tied closely to the causal graph and additive heuristics, this method selects optimal contexts for condition evaluations. It formulates as a shortest-path problem on a graph where nodes, representing problem atoms, are contextually labeled. This approach adeptly captures side effects, leading to enhanced heuristic evaluations. \cite{helmert2008unifying}

\subsubsection{Max-cost Heuristic} evaluates the cost of achieving a set of goals by recursively determining the maximum cost for each individual goal. For each goal, the heuristic computes the minimum cost over all actions that can produce the goal, considering the action's cost and the max-cost of its preconditions. \cite{ghallab2016automated} 

\subsubsection{Goal Count Heuristic} estimates the number of unsatisfied goals in a state, prioritizing states with fewer unsatisfied goals. This method is useful in problems with multiple goals, such as game playing and planning, and can improve the efficiency of finding a solution.

\subsubsection{LM-cost partitioning Heuristic} is a technique that distributes the costs of operators among the landmarks they achieve. By doing so, it ensures that the heuristic value remains admissible. In essence, the cost of each operator is divided among the landmarks it helps achieve, ensuring that the sum of the heuristic values of these landmarks does not exceed the actual cost. This approach allows for a more informed estimate of the cost to reach the goal by leveraging the structure of the problem's landmarks and the operators' costs. \cite{karpas2009cost}

\subsubsection{FF Heuristic} derived from the FF planning system, estimates the distance to the goal using a relaxed plan. In this relaxed scenario, negative action effects are ignored, and actions can be parallelized. By counting actions in the relaxed plan, the heuristic provides a lower bound on steps needed for the actual solution, guiding the search efficiently. \cite{hoffmann2001ff}

\subsubsection{Merge and Shrink Heuristic} is a technique for generating lower bounds in factored state spaces. Originating from model checking of automata networks, it offers a more general class of abstractions than traditional pattern databases. By merging and shrinking states, it can represent a broader range of abstractions, potentially more compactly. In essence, merge-and-shrink provides a balance between abstraction size and heuristic accuracy, making it a valuable tool for heuristic search in complex domains \cite{helmert2014merge}. In the Merge and Shrink (M\&S) heuristic we use bisimulation as shrinking strategy \cite{nissim2011computing}, strongly connected components as merging strategy \cite{sievers2016analysis}, and exact label reduction \cite{sievers2014generalized}. We limit the abstractions to 50,000 states.
        
\subsubsection{Pattern Database Heuristics}
The key step in using Pattern Database (PDBs) heuristics is selecting appropriate patterns for the problem at hand. \citet{korf1997finding} specified two sets of patterns for solving the Rubik's Cube. We evaluate two settings of PDBs:

{\em \textbf{Max Manual PDB:}} Inspired by Korf's patterns, \citet{buchner2022comparison} have considered 2 patterns for the corner cube pieces and 3 patterns for the edge cube pieces resulting in 4 variables for each pattern. We have considered these patterns for the evaluation of PDDL and SAS+ models.

{\em \textbf{Max Systematic PDB:}} This configuration systematically generates all
interesting patterns up to a certain size \cite{pommerening2013getting}. A pattern size of 3 has been considered for this evaluation in the interest of memory constraints.

\subsection{Experimental Setup}
To compare the performance of our RC PDDL model with the existing literature work, we have used the benchmark problem test set presented by \citet{buchner2022comparison}. In the benchmark test set, the problem tasks have been generated using 18 actions of RC - 12 actions correspond to 90-degree rotations of each face in clockwise and anti-clockwise directions, and the additional six actions are 180-degree rotation (suffix '2') on each face. The problem test set consists of 200 problems of varying difficulties. We have considered the scramble sequences provided to generate the respective PDDL versions of the problems. Additionally, we have generated our own test set of 200 RC problems considering only 12 actions. The problem generator starts from the goal state of RC and applies \textit{n} arbitrary actions from the list of 12 available actions. For every value of \textit{n}, ten unique random problem states are generated. The value of \textit{n} is between 1 and 20. The upper limit of 20 is chosen because the authors in \cite{rokicki2008twenty} state that all the RC problem instances can be solved with at-most 20 moves. It has been considered that every consecutive rotation corresponds to a different face of the RC, as such rotations can not be combined into a single rotation. 


The main difference between the two datasets \textit{d1} and \textit{d2} is that a 180-degree turn (half-turn) is considered as two actions in generating dataset \textit{d1}, while it has been considered as a single action in generating dataset \textit{d2}. The reason for evaluating two different datasets is that we wanted to capture the performance difference between the two PDDL models \textit{m1} (12 actions) and \textit{m2} (18 actions) in accordance with the difference in the branching factor. The PDDL model \textit{m2} and SAS+ model have similar branching factors.

To evaluate the RC PDDL model, we have used Scorpion planner \cite{seipp2020saturated}, which is an extension of Fast-Downward planner \cite{helmert2006fast}. Scorpion planner contains the implementation for PDBs that support conditional effects modeled in the domain file. We perform A* searches with each heuristic mentioned above on the test sets and the two PDDL models. We bound the A* search with an overall time limit of 30 minutes and a memory limit of 3.5GB. This constraint is the same for the abstraction heuristics as well, despite the fact that these heuristics require significant time for preprocessing and generating abstractions prior to the start of the search.


\begin{figure*}[!t]
    \centering
    \begin{subfigure}[t]{\textwidth}
        \centering
        \includegraphics[scale=0.5]{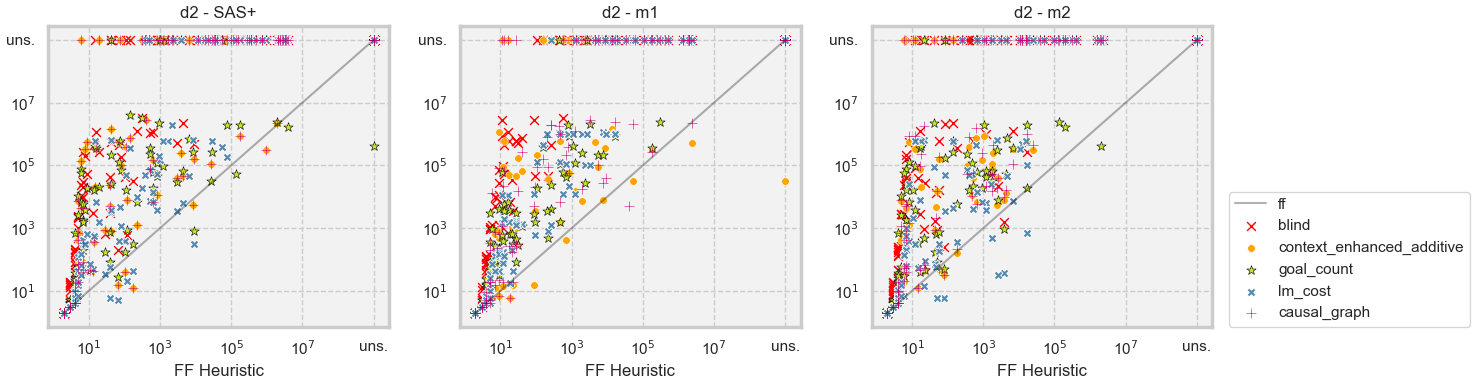}
        \caption{Non-abstraction heuristics comparison evaluated using FastDownward Planner. FF heuristic has the least state expansion trend across the representations.}
    \end{subfigure}

    \begin{subfigure}[t]{\textwidth}
        \centering
        \includegraphics[scale=0.48]{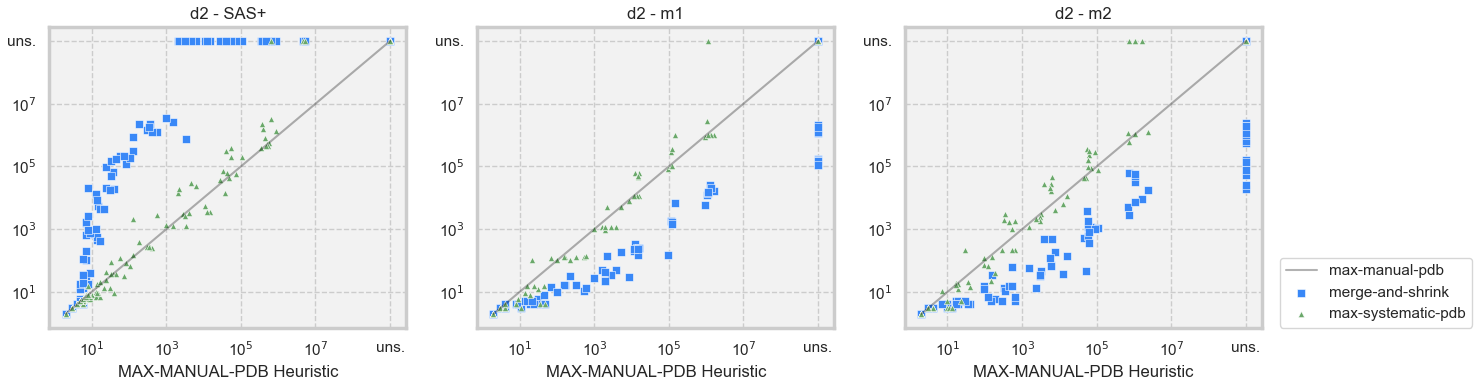}
        \caption{Abstraction heuristics comparison evaluated using Scorpion Planner. M\&S heuristic has lesser state expansion trend in PDDL than SAS+.}
    \end{subfigure}
    \caption{Comparison of the number of states expanded for dataset \textit{d2}.}
    \label{FIG: states_expanded}
\end{figure*}
\section{Result Analysis}

\begin{figure*}[!t]
    \centering
    \begin{subfigure}[t]{\textwidth}
        \centering
        \includegraphics[scale=0.5]{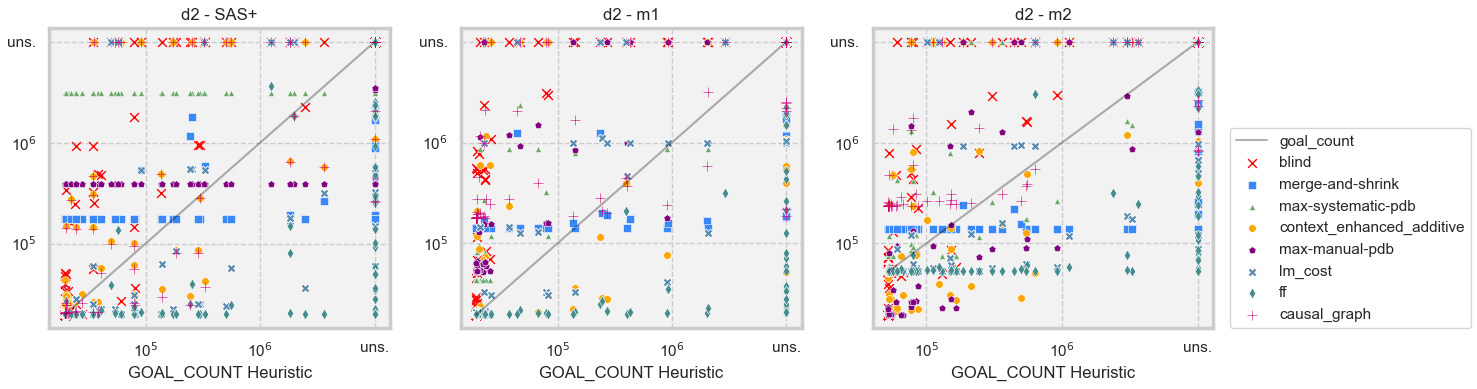}
        \caption{Memory usage comparison (unit in KB). See text for implications.}
    \end{subfigure}

    \begin{subfigure}[t]{\textwidth}
        \centering
        \includegraphics[scale=0.5]{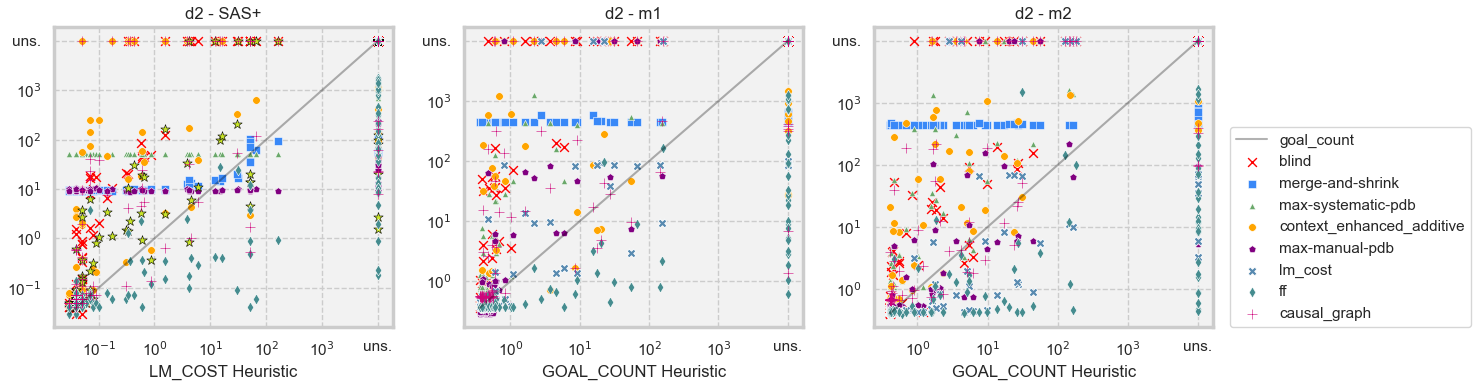}
        \caption{Run time comparison (unit in seconds). See text for implications.}
    \end{subfigure}
    \caption{Comparison of memory usage and runtime for different heuristics and models for dataset \textit{d2} evaluated using FastDownward Planning system.}
    \label{FIG: memory_time}
\end{figure*}


\subsection{Comparision of Heuristics}

We conducted an empirical evaluation of the performance of two different PDDL models (m1 and m2), each with varying numbers of modeled actions, on two test datasets (d1 and d2). Furthermore, we compared the efficacy of various heuristics on the SAS+ dataset provided by \citet{buchner2022comparison}. We also evaluated the test datasets using DeepCubeA \cite{agostinelli2019solving}, a state-of-the-art domain-independent RC solver that leverages a combination of deep reinforcement learning and search algorithms. Our results show that DeepCubeA was able to solve all the problems in both datasets, albeit with a lower percentage of optimal plans. We provide a detailed explanation of plan optimality in the subsequent section. Table \ref{tab:exp-results} presents the experimental results, including the total number of problems solved and the percentage of optimal plans generated for each configuration tested. Our findings offer valuable insights into the efficiency and effectiveness of the models, heuristics, and representations employed for solving RC problems. \\
1. It has been observed that abstraction heuristics are sensitive to problem representation and exhibit poorer performance in PDDL compared to SAS+ representation. \\
2. Interestingly, the FF heuristic, which is a non-abstraction heuristic, has been found to perform equally (in SAS+) or better than the state-of-the-art PDB heuristic with Korf's patterns in the case of PDDL representation. \\
3. The CG and CEA heuristics may not be effective for solving Rubik's Cube, a puzzle-solving domain with no modeled preconditions. The complex nature of the domain and large branching factor makes it challenging to construct an accurate causal graph for the CG heuristic, and the lack of contextual information renders CEA heuristic ineffective.\\
4. Modeling a domain with all possible actions leads to an increase in the number of optimal plans.

In this paper, we provide plots that compare the states expanded, runtime, and memory usage for the dataset \textit{d2} using both PDDL and SAS+ representations. We have chosen to focus on the performance of dataset \textit{d2} because it can be compared with the SAS+ dataset provided in the study by \citealt{buchner2022comparison} as they have similar branching factor. Comparable figures for dataset \textit{d1} are available in the supplementary material, and the conclusions reached in the paper are consistent with those results. Additionally, we include supplementary plots that depict the number of states expanded, memory usage, and runtime comparisons against all other heuristics for both datasets and all models.

When assessing the efficacy of planning-based solvers in terms of their heuristics and representations, our findings indicate that no planner configuration was able to solve problems with optimal plan lengths exceeding 13 steps. However, DeepCubeA was capable of solving problems up to 26 steps in length. In terms of the number of problems solved, the FF heuristic is the best performing across all models in both datasets. The M\&S abstraction heuristic is the second-best performing heuristic in PDDL representation, but this is not the case for SAS+ representation. In fact, PDBs performed much better in the SAS+ representation than in the PDDL representation and were equally as effective as the FF heuristic. This can also be inferred from the states expansion trend of the abstraction heuristics shown in Figure \ref{FIG: states_expanded}(b).
The reason why pattern databases performed better in SAS+ representation than in PDDL representation and were equally effective as the FF heuristic is due to the greater expressiveness of SAS+ models, which allow for more efficient and compact representations of problems. Pattern databases are better able to capture the structure and relationships of problems in SAS+ representation, leading to more accurate and effective heuristics. However, preprocessing time for pattern databases can be longer in SAS+ representation because the language is more explicit and requires the computation of more states to generate the pattern. This is evident from the runtime comparison plot shown in Figure \ref{FIG: memory_time}(b).

\subsubsection{Comparison based on states expanded:}
Figure \ref{FIG: states_expanded} illustrates the number of states expanded in the A\textsuperscript{*} search algorithm. Specifically, Figure \ref{FIG: states_expanded}(a) compares the performance of various non-abstract heuristics against the FF heuristic, which was found to be the best-performing heuristic for solving the given set of problems. The diagonal line in the plot represents the performance of the heuristics if they were to perform equally well as the FF heuristic. Heuristics that perform better than the FF heuristic would appear below the diagonal, while heuristics that perform worse would appear above it. On the plot in Figure \ref{FIG: states_expanded}(a), the unsolved points on the \textit{y}-axis represent the set of problems that were unable to be solved by the other heuristics, while the FF heuristic was able to solve them. Conversely, the unsolved points on the \textit{x}-axis represent the set of problems that the FF heuristic was unable to solve, while the other heuristics were able to solve them. This applies to all other plots provided in the paper. As seen in the plot, all non-abstract heuristics performed worse than the FF heuristic as they lie on the left side of the diagonal. This indicates that the number of states expanded by these heuristics is higher than that of the FF heuristic. This finding provides an explanation as to why the other heuristics performed poorly within the given time and memory constraints when compared to the FF heuristic.

Figure \ref{FIG: states_expanded}(b) displays the trend of state expansion for abstraction heuristics compared to PDB-Man. The reason for selecting PDB-Man was to allow for an interesting comparison between Merge-and-Shrink (M\&S) and Pattern Database (PDBs) heuristics across different problem representations. It is observed that the state expansion trend for PDB-Man and PDB-Sys is identical, while M\&S performance varies depending on the representation used. Specifically, M\&S is found to expand more states than PDBs in the case of the SAS+ model, while in PDDL models, this is not the case. In fact, M\&S performs better in PDDL models, where it expands fewer states than PDBs. These results suggest that the choice of abstraction heuristic can have a significant impact on search algorithm performance, depending on the problem representation. Different abstraction heuristics may be better suited for different problem representations, emphasizing the need for careful evaluation to determine the most effective heuristic for each representation.

\subsubsection{Runtime and memory usage comparison:}
Figure \ref{FIG: memory_time} presents a comparison of the runtime and memory usage of all considered heuristics. Figure \ref{FIG: memory_time}(a) presents a comparison of the memory usage for all heuristics, plotted against the GC heuristic, which exhibits an evenly distributed memory usage pattern for problems with different difficulties among the considered heuristics. Similarly, Figure \ref{FIG: memory_time}(b) displays the runtime comparison for the considered heuristics, plotted against the LM-Cost heuristic for the SAS+ model and GC heuristic for PDDL models as they exhibit a comparatively better distribution of runtime across the problems of varying difficulty in the respective representations.

The following observations have been made from the comparison of runtime and memory usage of different heuristics and problem representations shown in Figure \ref{FIG: memory_time}.

    \noindent 1. In the case of SAS+ models, the preprocessing time of pattern database (PDB) heuristics is higher and remains constant even for trivial tasks. However, this is not the case for PDDL models, where the preprocessing time does not exhibit a constant trend and takes significantly less time for trivial tasks. This observation highlights the impact of the problem representation on the preprocessing time of PDB heuristics.

    \noindent 2. The M\&S heuristic exhibits a higher runtime in PDDL models compared to SAS+ models, but the memory usage pattern remains similar across the representations. This is intriguing given that the bytes required to represent a single state are higher in PDDL models than in SAS+ models. The state expansion plots provided in Figure \ref{FIG: states_expanded} further support this observation, showing that M\&S heuristic expands more states in SAS+ representation than in PDDL.
    
    \noindent 3. In SAS+ representation, the preprocessing time of both M\&S and PDB-Man heuristics is similar. However, as the problem complexity increases, M\&S heuristic exhibits a higher search time compared to PDB-Man. This trend is evident in Figure \ref{FIG: memory_time}(b), where the constant line starts to ascend earlier for M\&S than for PDB-Man. These results suggest that while both heuristics may be suitable for simple problem instances, PDB-Man may offer better performance for more complex instances in SAS+ representation.
    
    \noindent 4. The runtime performance of the PDB-Sys heuristic is comparable to the blind heuristic, as both exhibit poor performance. This is supported by the runtime comparison plotted against PDB-Sys in the supplementary material.

    \noindent 5. In the case of PDDL, as the number of actions increases, the memory usage pattern for trivial tasks is found to be lesser for the PDB-Man heuristic. This suggests that PDB-Man is able to use its precomputed pattern database more effectively as the size of the planning problem increases. In contrast, for the FF heuristic, the memory usage pattern is found to be the reverse, indicating that as the size of the problem increases we find that there is an increase in the memory usage pattern.
    
    \noindent 6.  FF heuristic is the most efficient heuristic comparatively in terms of both runtime and memory usage across the representations. This explains the fact that the FF heuristic was able to solve the highest number of problems within the given time and memory budget.

    \noindent 7. For both SAS+ and PDDL representations, the runtime pattern of the causal graph and context-enhanced additive heuristics are similar. 
\subsection{Plan Optimality Analysis}

Table \ref{tab:exp-results} presents the performance of different planner configurations with various problem representations, including the number of problems solved and the percentage of optimal plans generated. To assess the optimality of the generated plans, we juxtaposed them with optimal plans produced by an optimal solver. This solver was developed by Michael Reid, drawing inspiration from methodologies pioneered by Herbert Kociemba \cite{kociemba2006cube}. The optimal solver can be accessed at \footnote{\url{https://www.cflmath.com/~reid/Rubik/optimal_solver.html}}. The table also includes the performance of DeepCubeA on the considered datasets. It was found that DeepCubeA was able to solve all the problems in both datasets. However, the percentage of optimal plans generated by DeepCubeA are \textit{94.5\%} and \textit{78.5\%} for datasets \textit{d1} and \textit{d2}, respectively. The poor performance of DeepCubeA on dataset \textit{d2} can be attributed to the fact that DeepCubeA was trained with only 12 RC actions in its modeling\footnote[5]{The performance of DeepCubeA to generate optimal plans may increase if 18 RC actions were considered in the training phase. We followed the documentation and data provided in \cite{deepcubea-code} while training the model which has only 12 RC actions modeled.}, whereas the dataset has problem states generated considering 18 RC actions. Furthermore, the actions sequence generated by DeepCubeA for a problem instance in dataset \textit{d2} can be further simplified when considering the 18 RC actions set. For instance, if the RC is shuffled with action \textit{F2}, which is a 180-degree turn on the front face, DeepCubeA generates a plan of sequence (\textit{F}, \textit{F}), where these two actions can be combined into a single action \textit{F2}. This also explains why DeepCubeA has a higher percentage of optimal plans for dataset \textit{d1}. This is the same reason for the lower percentage of optimal plans in the case of dataset \textit{d2} tested with PDDL model \textit{m1}. We find that the percentage of optimal plans for RC increases as the number of actions modeled in the PDDL increases.

\section{Discussion and Conclusion}

    





In this study, we conducted an extensive comparison of planning-based and learning-based approaches to solve a complex combinatorial problem: the 3x3x3 RC. We evaluated the effectiveness of existing SAS+ and custom representations for RC, and introduced the first PDDL representation. We examined the capabilities of different heuristics for various representational configurations in solving RC. Our results indicate that a symbolic planner can benefit from using SAS+ representation, which offers a more compact state representation that is approximately $\sim$75\% more memory efficient than PDDL. However, a specific planner configuration could only solve 61.50\% of RC problems with 100\% optimality. In contrast, the DeepCubeA learning-based approach was able to solve 100\% of the problems sub-optimally (worst-case 18\%). However, the current custom representation used by DeepCubeA may have an impact on the optimal plan generation as it lacks any semantics representing RC. Based on our experimental insights, we note that using SAS+ representation to encode RC problems for DeepCubeA and learning PDBs instead of its current weighted A* search, may improve plan optimality ratio. We also note that while traditional planners generate higher optimal plans, they are limited to solving only 0.0001\% of the $4.3x10^{19}$ states of RC \cite{rokicki2014diameter}. However, DeepCubeA is able to solve for all possible states of 3x3x3 RC. Our study highlights the potential of both automated and learning-based planners, and suggests a unified approach that can generalize to higher-dimensional RC configurations while preserving the solving capabilities of a learned approach and the optimality of a traditional planner.

\bibliography{references/references}

\end{document}


\begin{titlepage}
    \centering
    \vspace*{1cm}
    \Large
    \textbf{Supplementary material}

    \vspace{1cm}
    \begin{flushleft}
        \section{Introduction}
        In the following supplementary material we provide additional plots for comparison of the number of states expanded, runtime and memory usage comparisons plotted against every heuristic we considered in our paper. Also we provide these plots for dataset \textit{d1} and these are consistent with the conclusions mentioned in our paper. We have provided a description of these figures in Table \ref{tab: fig_description}
        \section{Hardware}
        We hve used two servers to run our experiments. One with 48-core nodes each hosting 2 V100 32G GPUs and 128GB of RAM. Another with 256-cores, eight A100 40GB GPUs and 1TB of RAM. The processor speed is 2.8 GHz.
    \end{flushleft}
    
    \begin{table}[!b]
        \centering
        \caption{Description of the figures}
        \label{tab: fig_description}
        \begin{tabular}{|l|p{15cm}|}
            \hline
            \textbf{Figure No.} & \textbf{Description} \\
            \hline
            Figure 1 & Comparison of the number of states expanded plotted against the Blind heuristic \\
            \hline
            Figure 2 & Comparison of the number of states expanded plotted against the Causal Graph heuristic \\
            \hline
            Figure 3 & Comparison of the number of states expanded plotted against the Context-enhanced Additive heuristic \\
            \hline
            Figure 4 & Comparison of the number of states expanded plotted against the Goal Count heuristic \\
            \hline
            Figure 5 & Comparison of the number of states expanded plotted against the LM-Cost heuristic \\
            \hline
            Figure 6 & Comparison of the number of states expanded plotted against the FF heuristic \\
            \hline
            Figure 7 & Comparison of the number of states expanded plotted against the Merge \& Shrink heuristic \\
            \hline
            Figure 8 & Comparison of the number of states expanded plotted against the Max-Manual PDB heuristic \\
            \hline
            Figure 9 & Comparison of the number of states expanded plotted against the Max-Systematic PDB heuristic \\
            \hline \hline
            Figure 10 & Comparison of memory usage plotted against the Blind heuristic \\
            \hline
            Figure 11 & Comparison of memory usage plotted against the Causal Graph heuristic \\
            \hline
            Figure 12 & Comparison of memory usage plotted against the Context-enhanced Additive heuristic \\
            \hline
            Figure 13 & Comparison of memory usage plotted against the Goal Count heuristic \\
            \hline
            Figure 14 & Comparison of memory usage plotted against the LM-Cost heuristic \\
            \hline
            Figure 15 & Comparison of memory usage plotted against the FF heuristic \\
            \hline
            Figure 16 & Comparison of memory usage plotted against the Merge \& Shrink heuristic \\
            \hline
            Figure 17 & Comparison of memory usage plotted against the Max-Manual PDB heuristic \\
            \hline
            Figure 18 & Comparison of memory usage plotted against the Max-Systematic PDB heuristic \\
            \hline \hline
            Figure 19 & Comparison of runtime plotted against the Blind heuristic \\
            \hline
            Figure 20 & Comparison of runtime plotted against the Causal Graph heuristic \\
            \hline
            Figure 21 & Comparison of runtime plotted against the Context-enhanced Additive heuristic \\
            \hline
            Figure 22 & Comparison of runtime plotted against the Goal Count heuristic \\
            \hline
            Figure 23 & Comparison of runtime plotted against the LM-Cost heuristic \\
            \hline
            Figure 24 & Comparison of runtime plotted against the FF heuristic \\
            \hline
            Figure 25 & Comparison of runtime plotted against the Merge \& Shrink heuristic \\
            \hline
            Figure 26 & Comparison of runtime plotted against the Max-Manual PDB heuristic \\
            \hline
            Figure 27 & Comparison of runtime plotted against the Max-Systematic PDB heuristic \\
            \hline
        \end{tabular}
    \end{table}

\end{titlepage}

\begin{figure*}[htbp]
  \centering
  \includegraphics[scale=0.5]{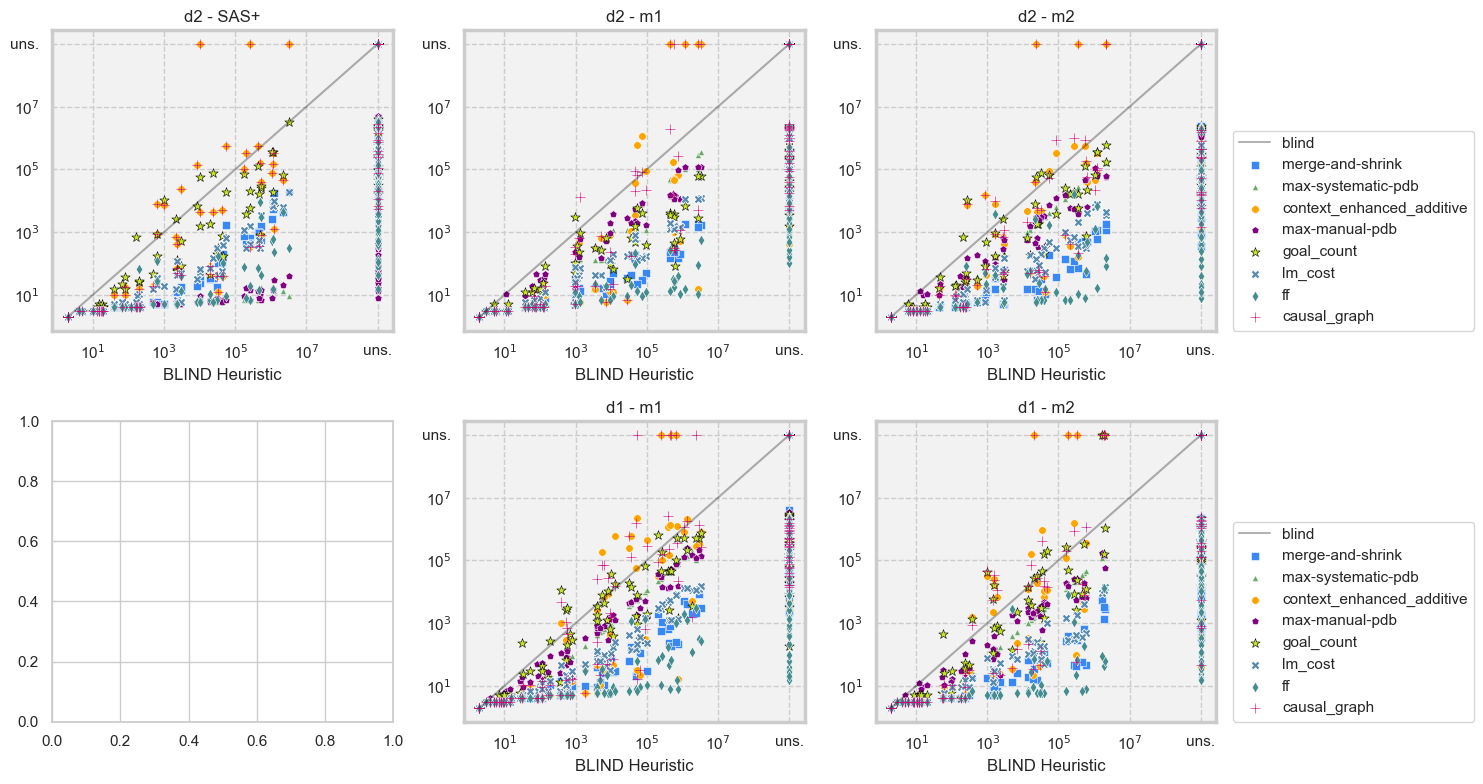}
  \caption{Comparison of the number of states expanded plotted against the blind heuristic}
\end{figure*}

\begin{figure*}[htbp]
  \centering
  \includegraphics[scale=0.5]{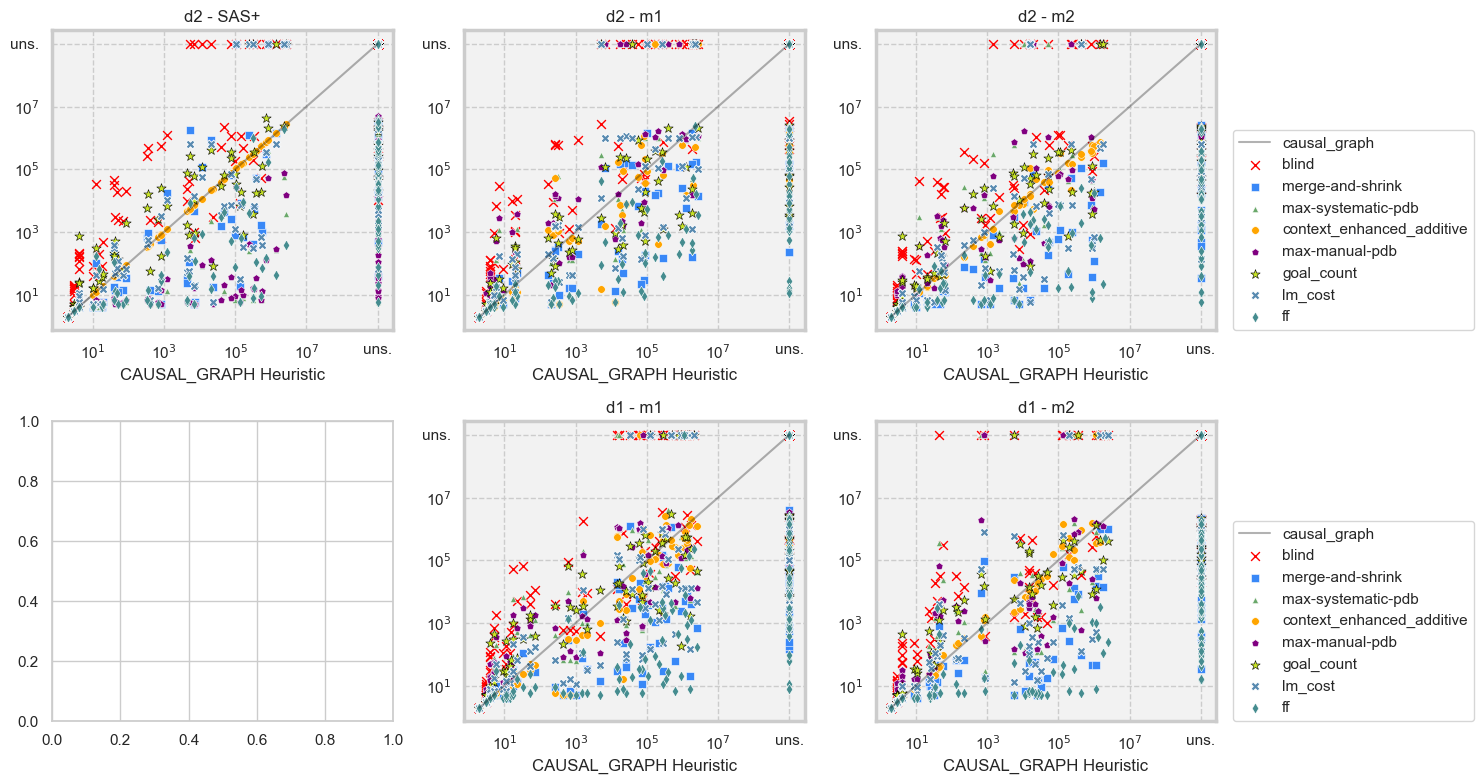}
  \caption{Comparison of the number of states expanded plotted against the causal graph heuristic}
\end{figure*}

\begin{figure*}[htbp]
  \centering
  \includegraphics[scale=0.5]{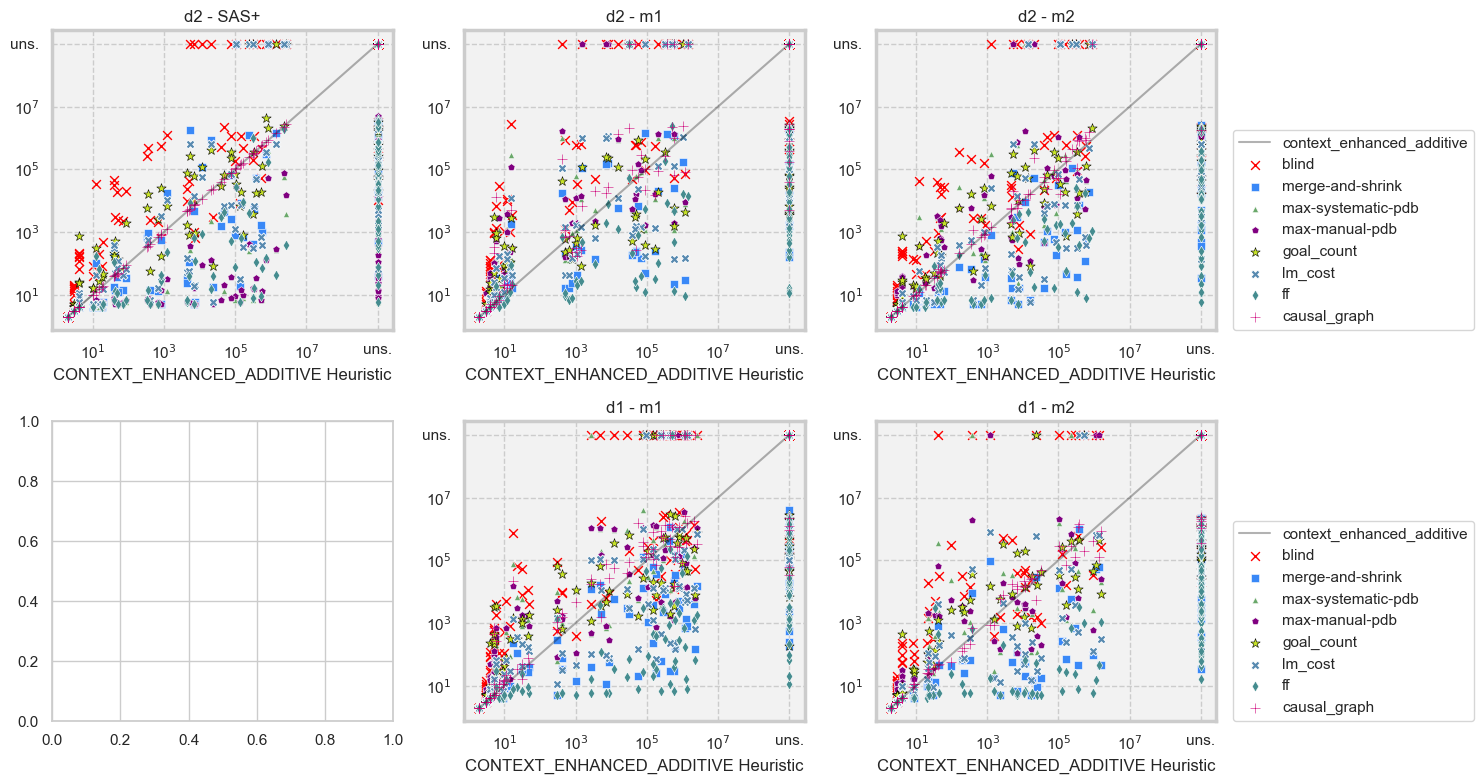}
  \caption{Comparison of the number of states expanded plotted against the context-enhanced additive heuristic}
\end{figure*}

\begin{figure*}[htbp]
  \centering
  \includegraphics[scale=0.5]{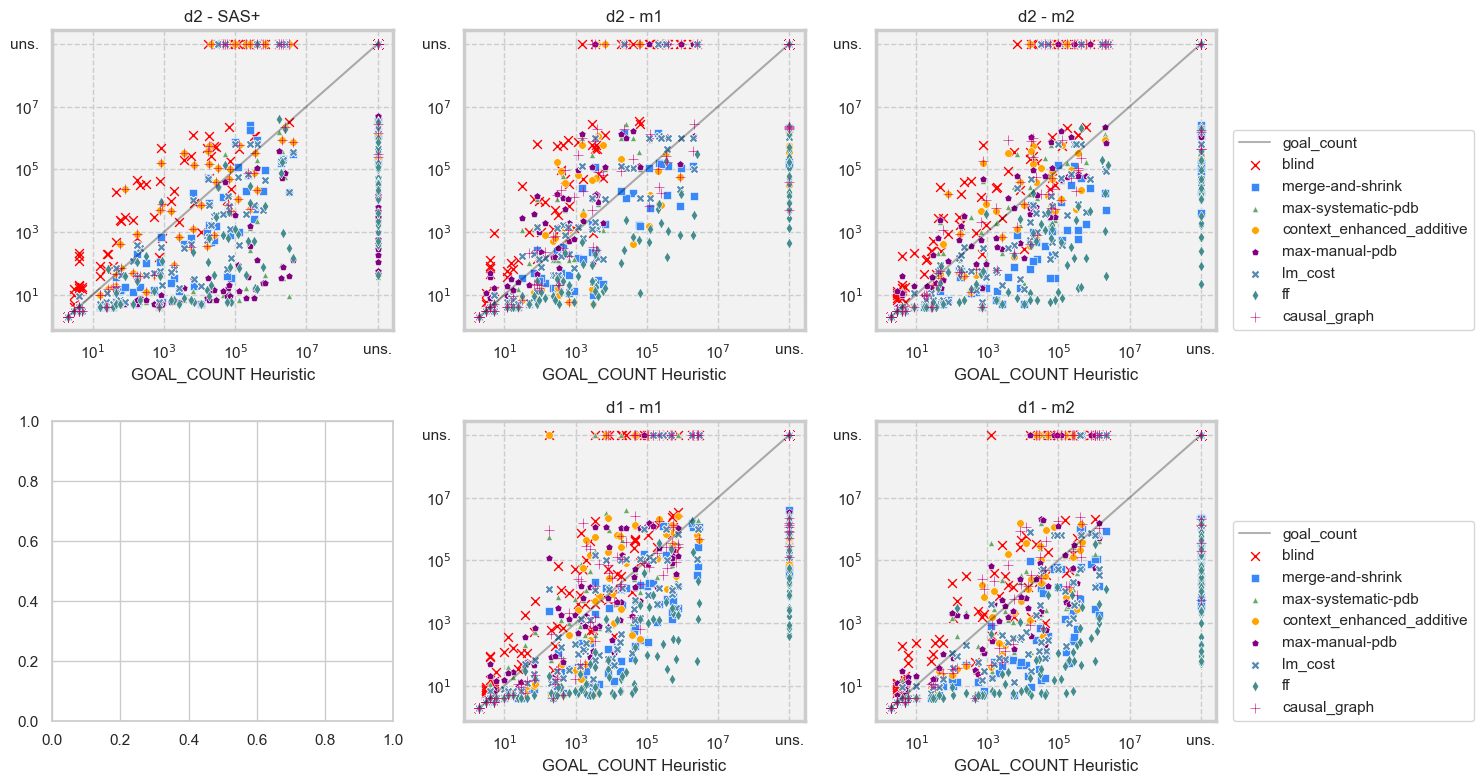}
  \caption{Comparison of the number of states expanded plotted against the goal count heuristic}
\end{figure*}

\begin{figure*}[htbp]
  \centering
  \includegraphics[scale=0.5]{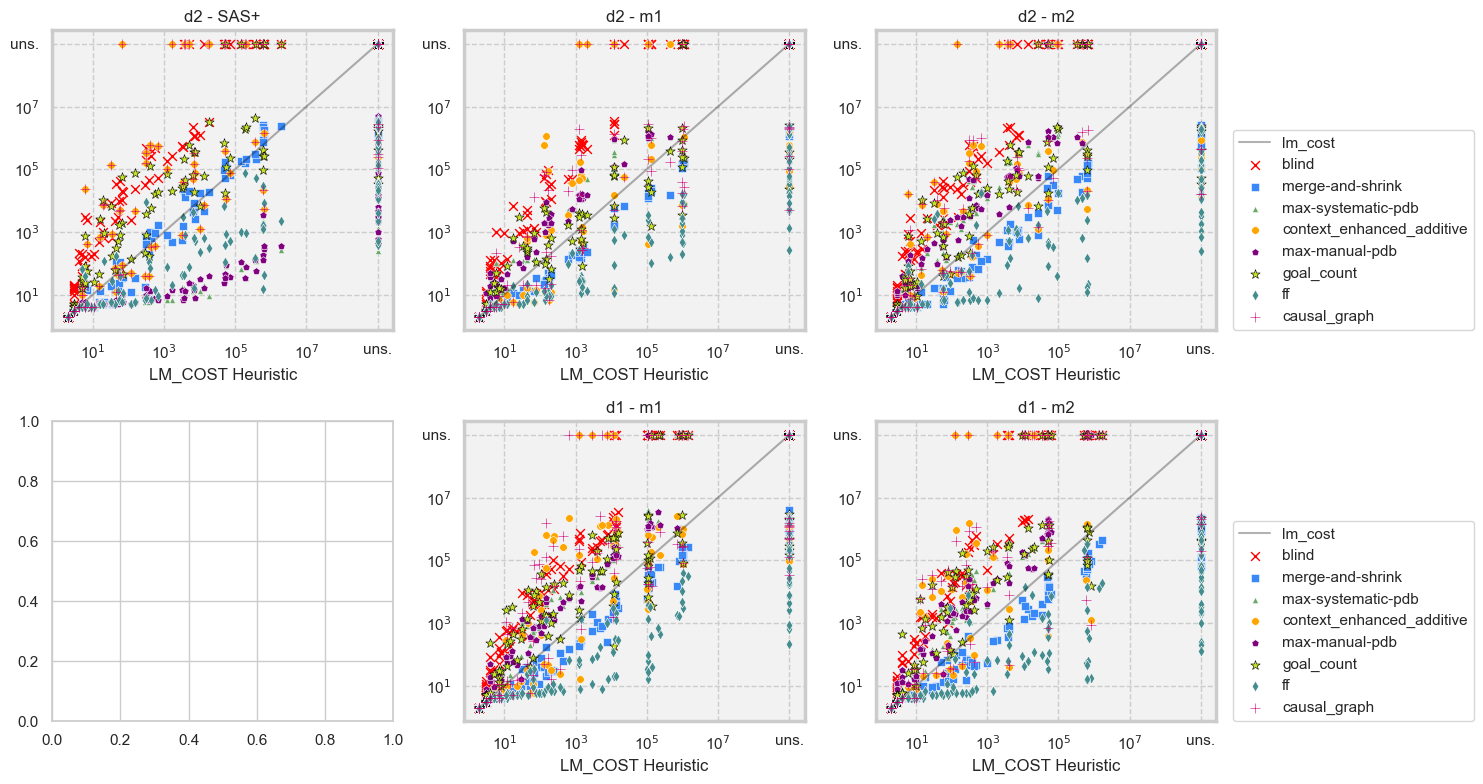}
  \caption{Comparison of the number of states expanded plotted against the LM-Cost heuristic}
\end{figure*}

\begin{figure*}[htbp]
  \centering
  \includegraphics[scale=0.5]{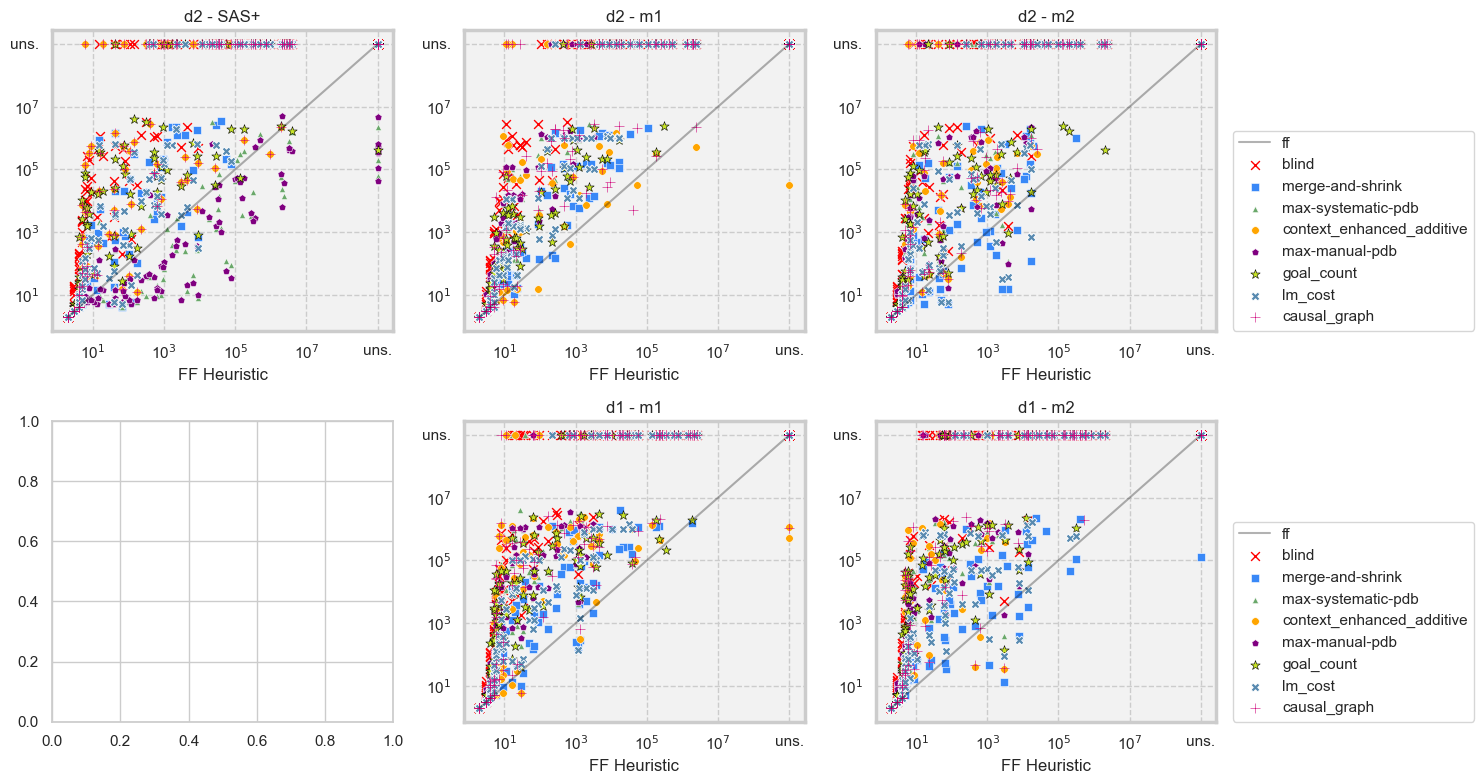}
  \caption{Comparison of the number of states expanded plotted against the FF heuristic}
\end{figure*}

\begin{figure*}[htbp]
  \centering
  \includegraphics[scale=0.5]{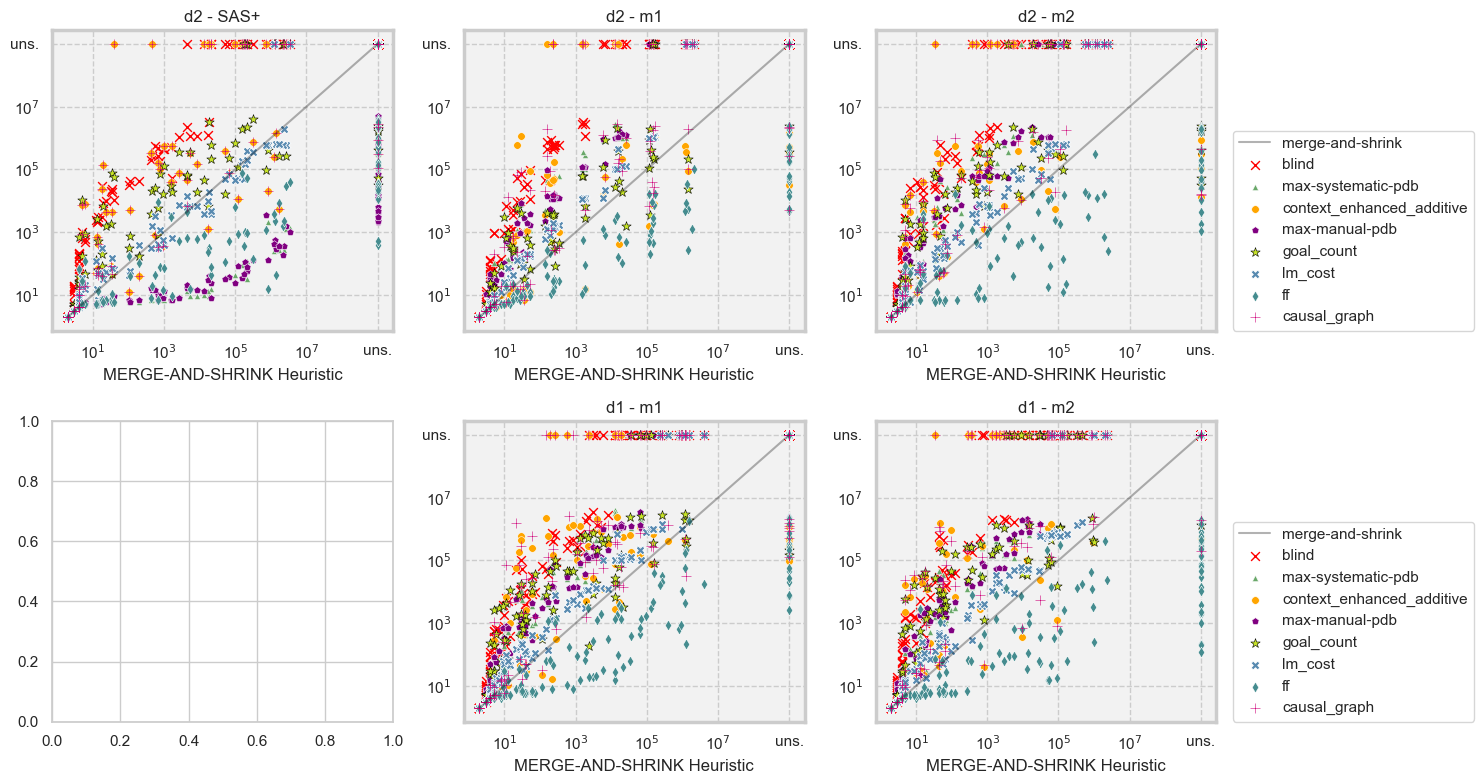}
  \caption{Comparison of the number of states expanded plotted against the Merge-and-Shrink heuristic}
\end{figure*}

\begin{figure*}[htbp]
  \centering
  \includegraphics[scale=0.5]{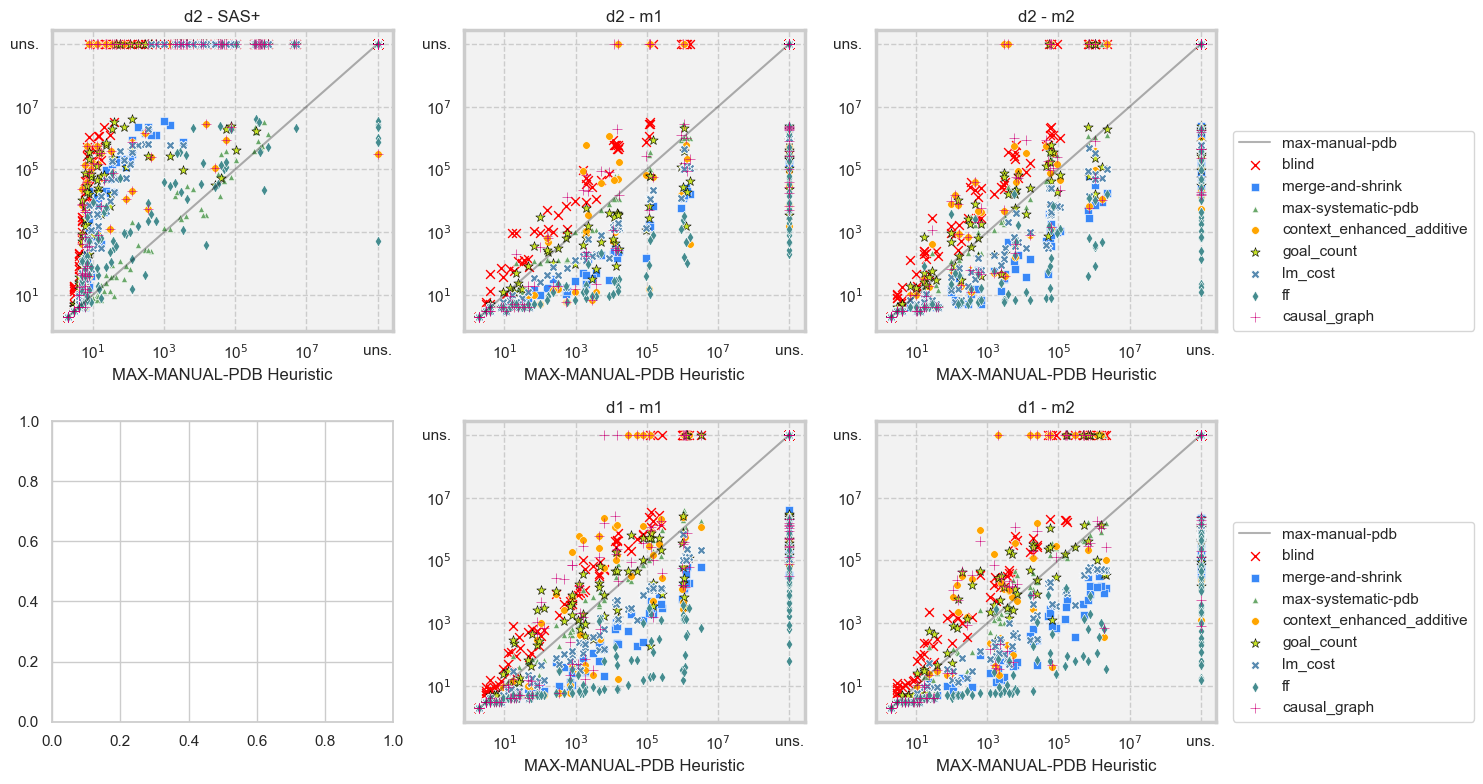}
  \caption{Comparison of the number of states expanded plotted against the Max-Manual-PDB heuristic}
\end{figure*}

\begin{figure*}[htbp]
  \centering
  \includegraphics[scale=0.5]{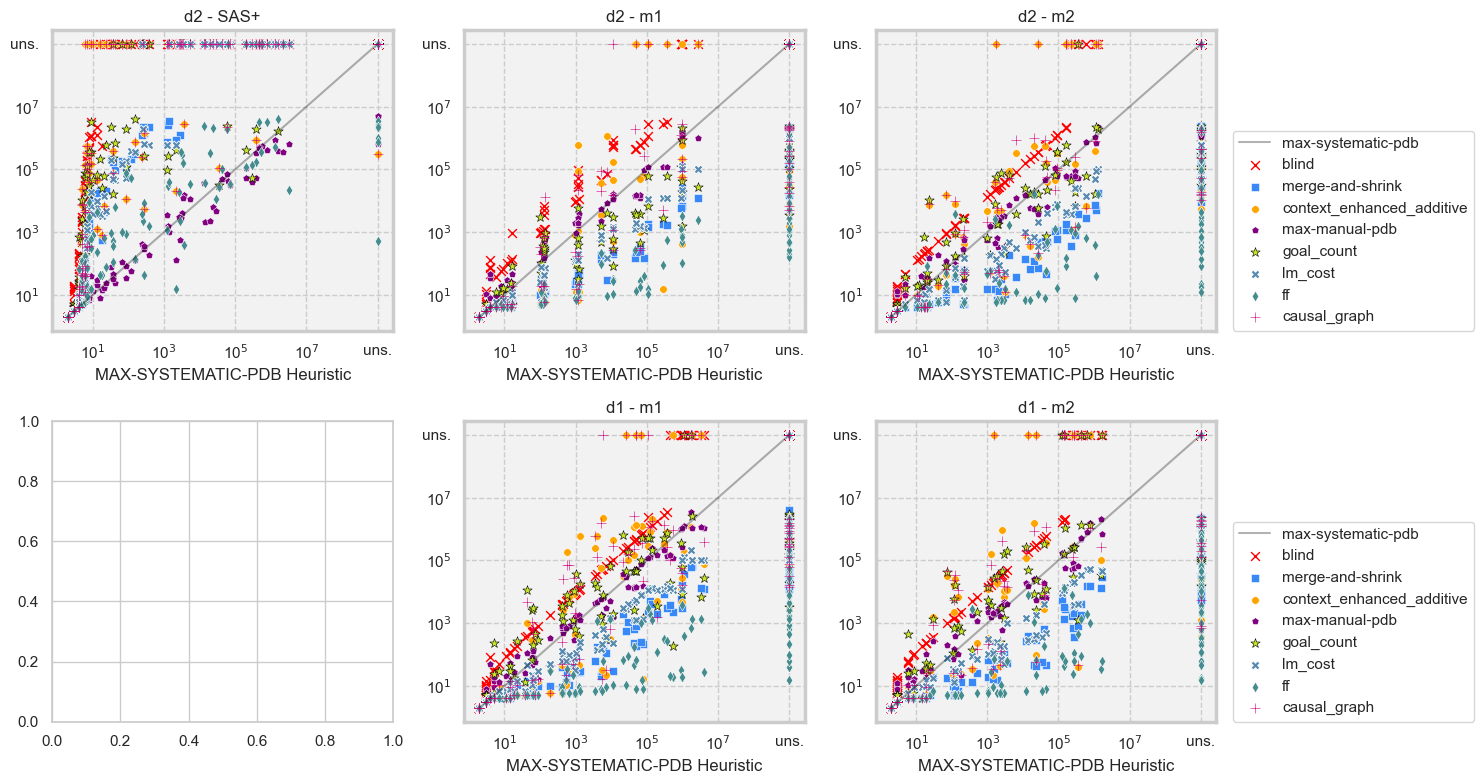}
  \caption{Comparison of the number of states expanded plotted against the Max-Systematic-PDB heuristic}
\end{figure*}


\begin{figure*}[htbp]
  \centering
  \includegraphics[scale=0.5]{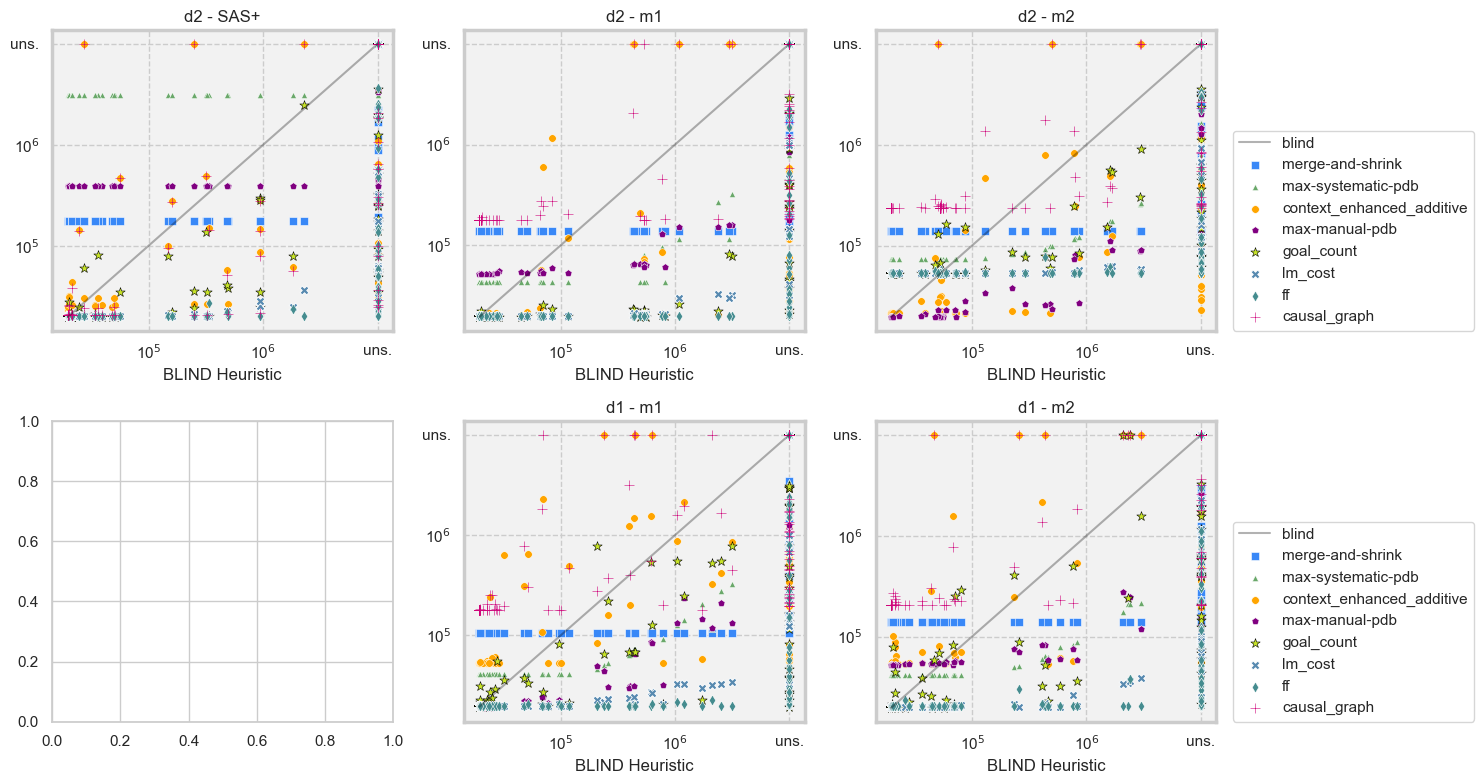}
  \caption{Comparison of memory usage plotted against the Blind heuristic}
\end{figure*}

\begin{figure*}[htbp]
  \centering
  \includegraphics[scale=0.5]{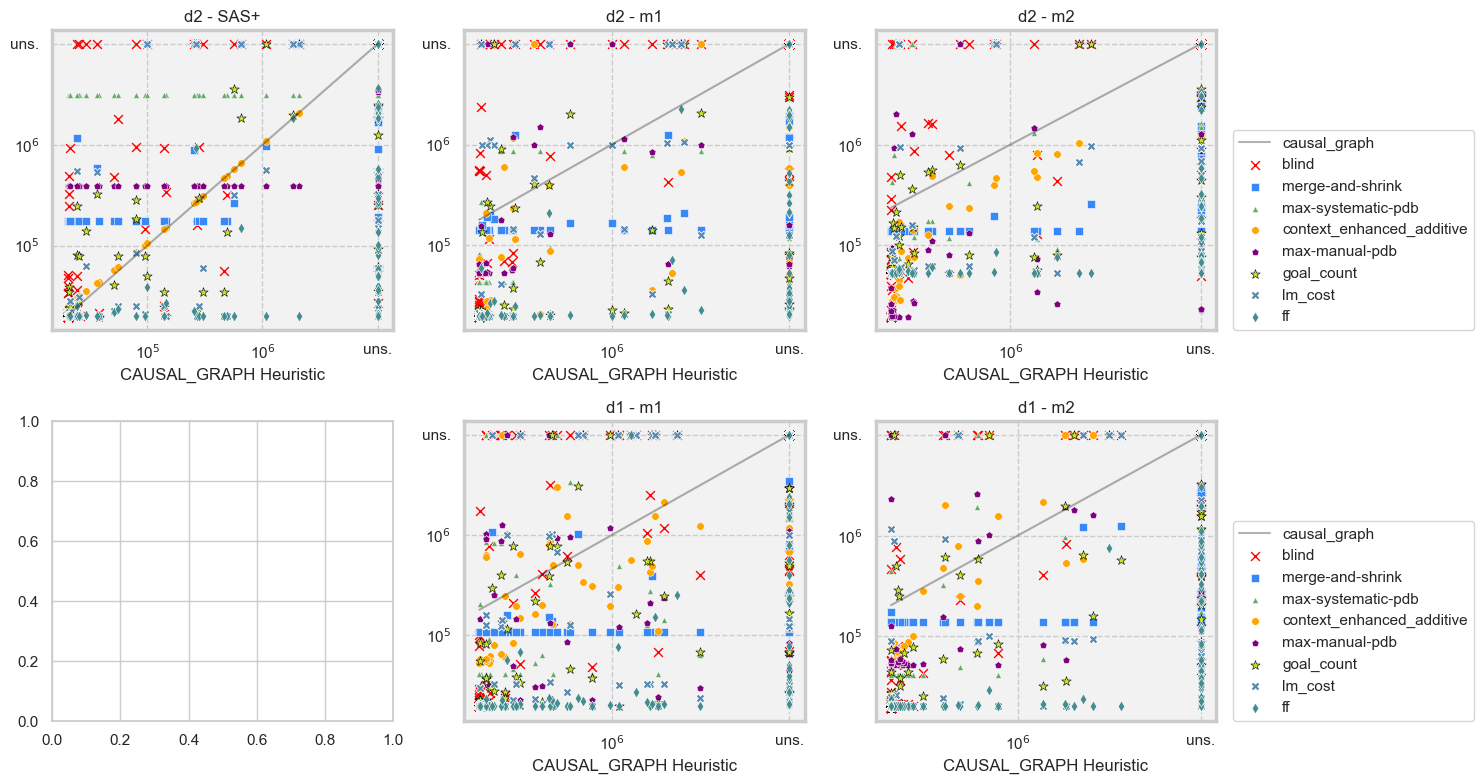}
  \caption{Comparison of memory usage plotted against the causal graph heuristic}
\end{figure*}

\begin{figure*}[htbp]
  \centering
  \includegraphics[scale=0.5]{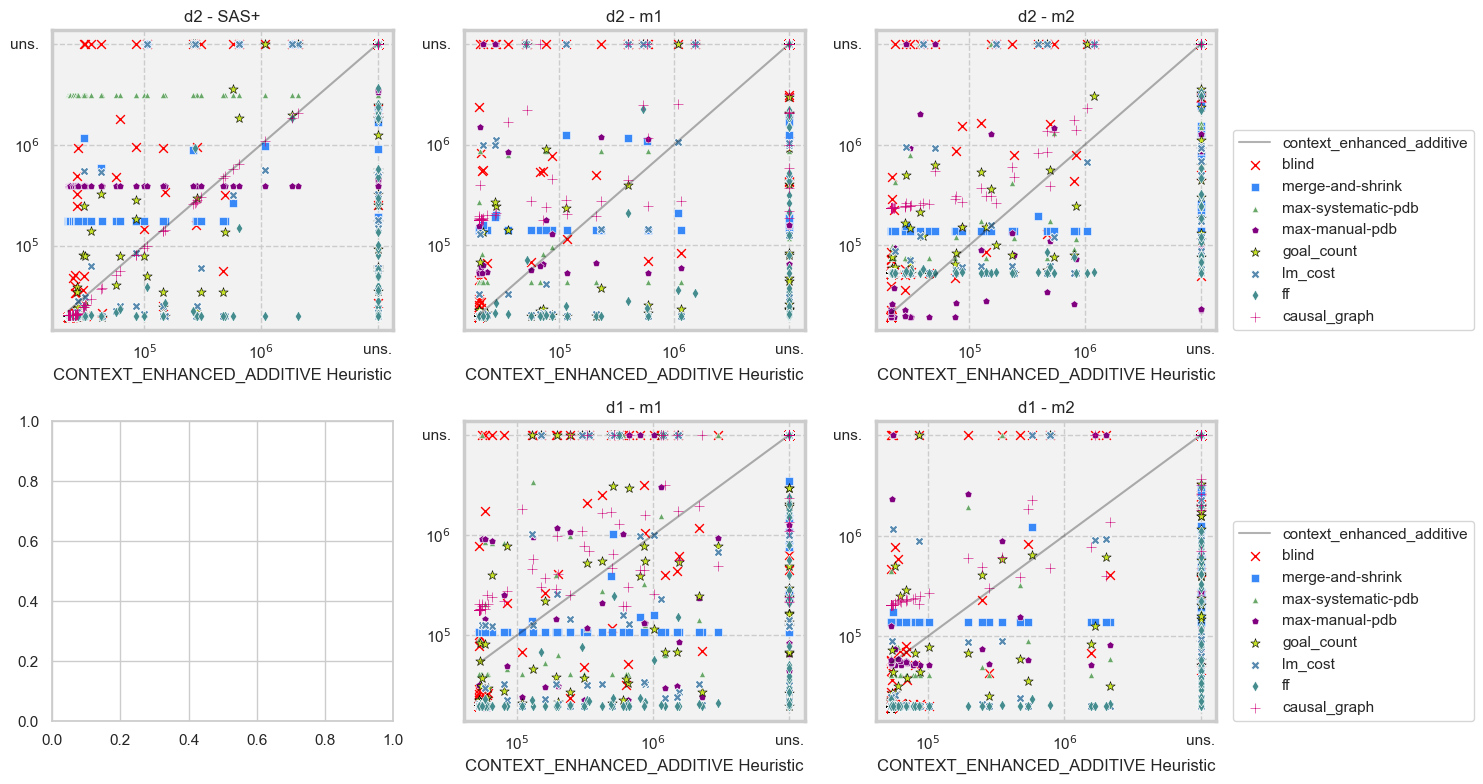}
  \caption{Comparison of memory usage plotted against the Context-Enhanced Additive heuristic}
\end{figure*}

\begin{figure*}[htbp]
  \centering
  \includegraphics[scale=0.5]{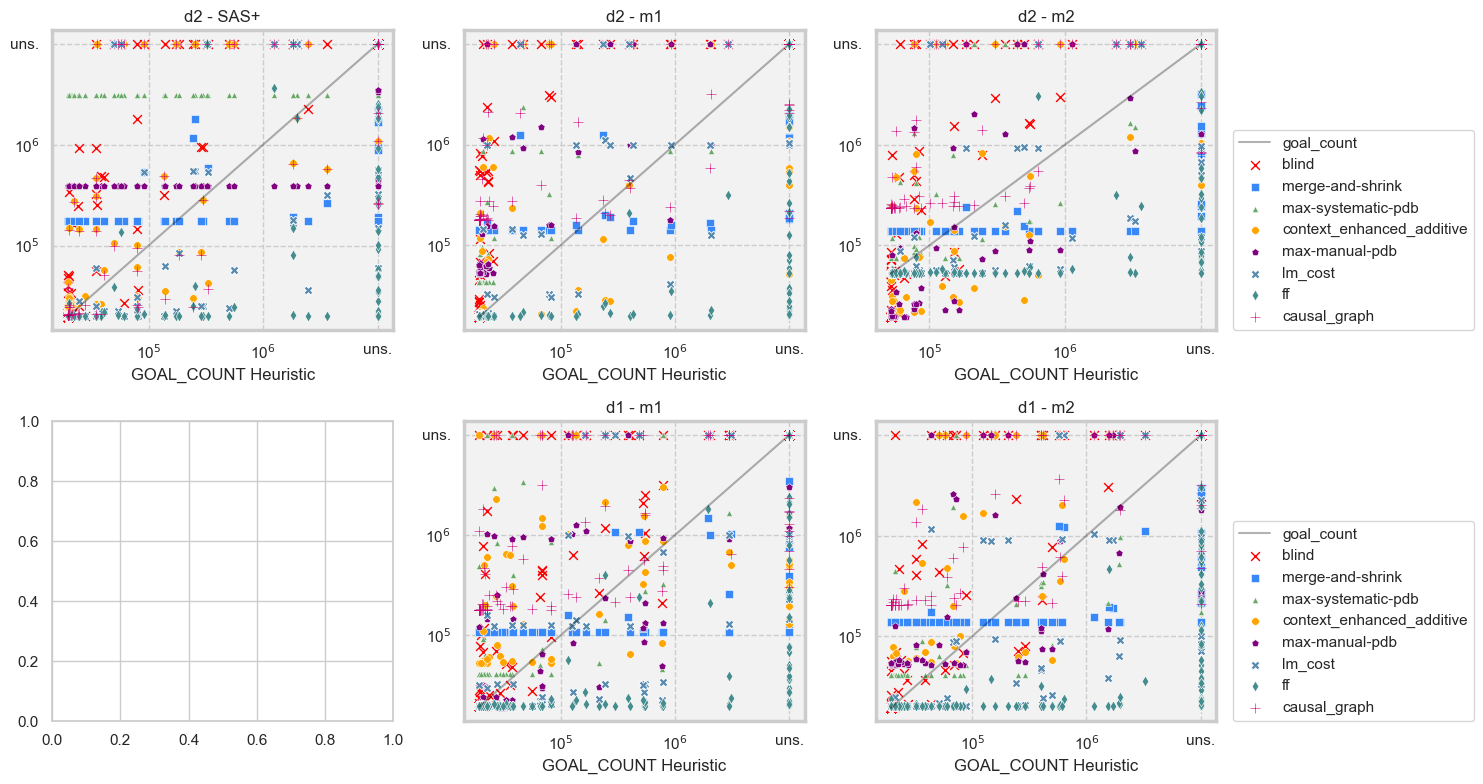}
  \caption{Comparison of memory usage plotted against the Goal Count heuristic}
\end{figure*}

\begin{figure*}[htbp]
  \centering
  \includegraphics[scale=0.5]{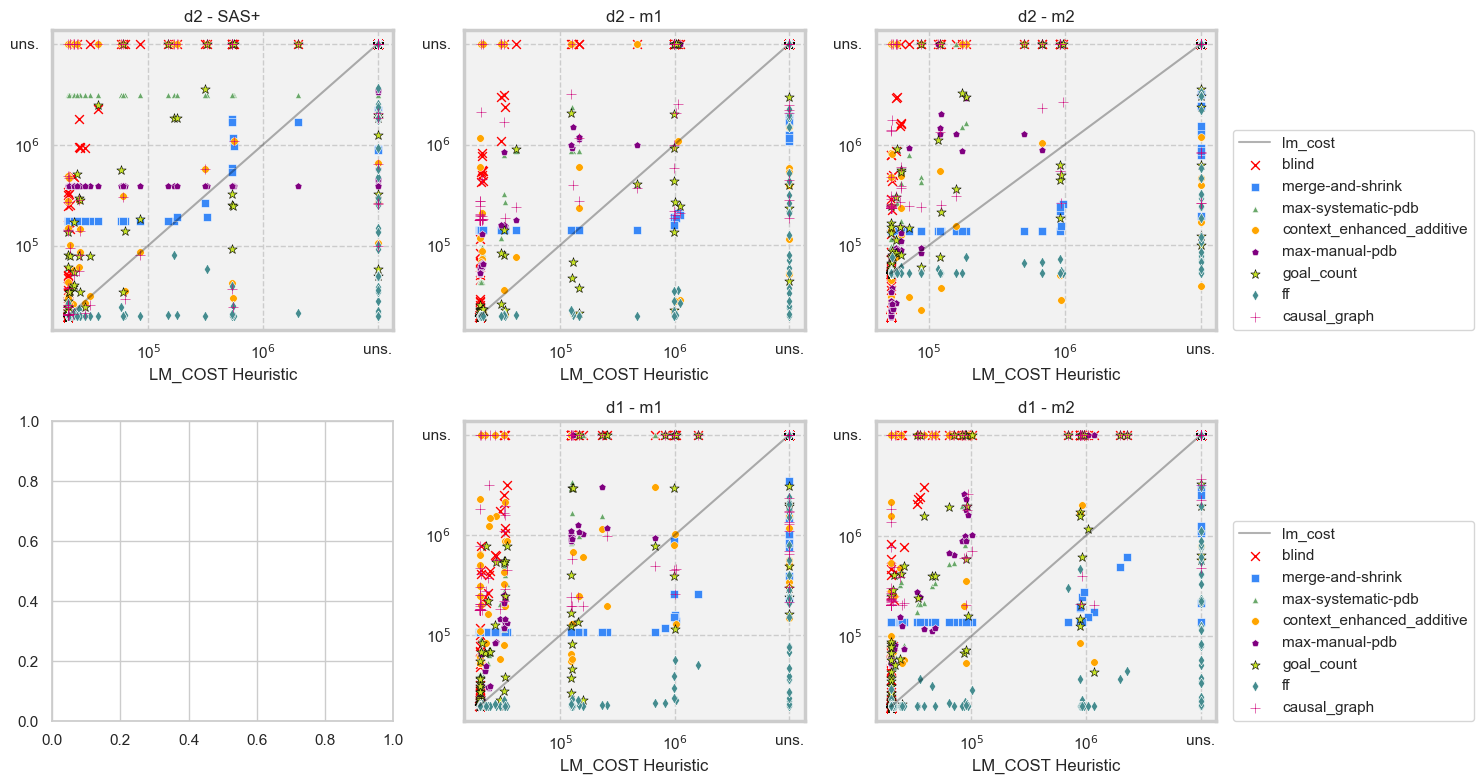}
  \caption{Comparison of memory usage plotted against the LM-Cost heuristic}
\end{figure*}

\begin{figure*}[htbp]
  \centering
  \includegraphics[scale=0.5]{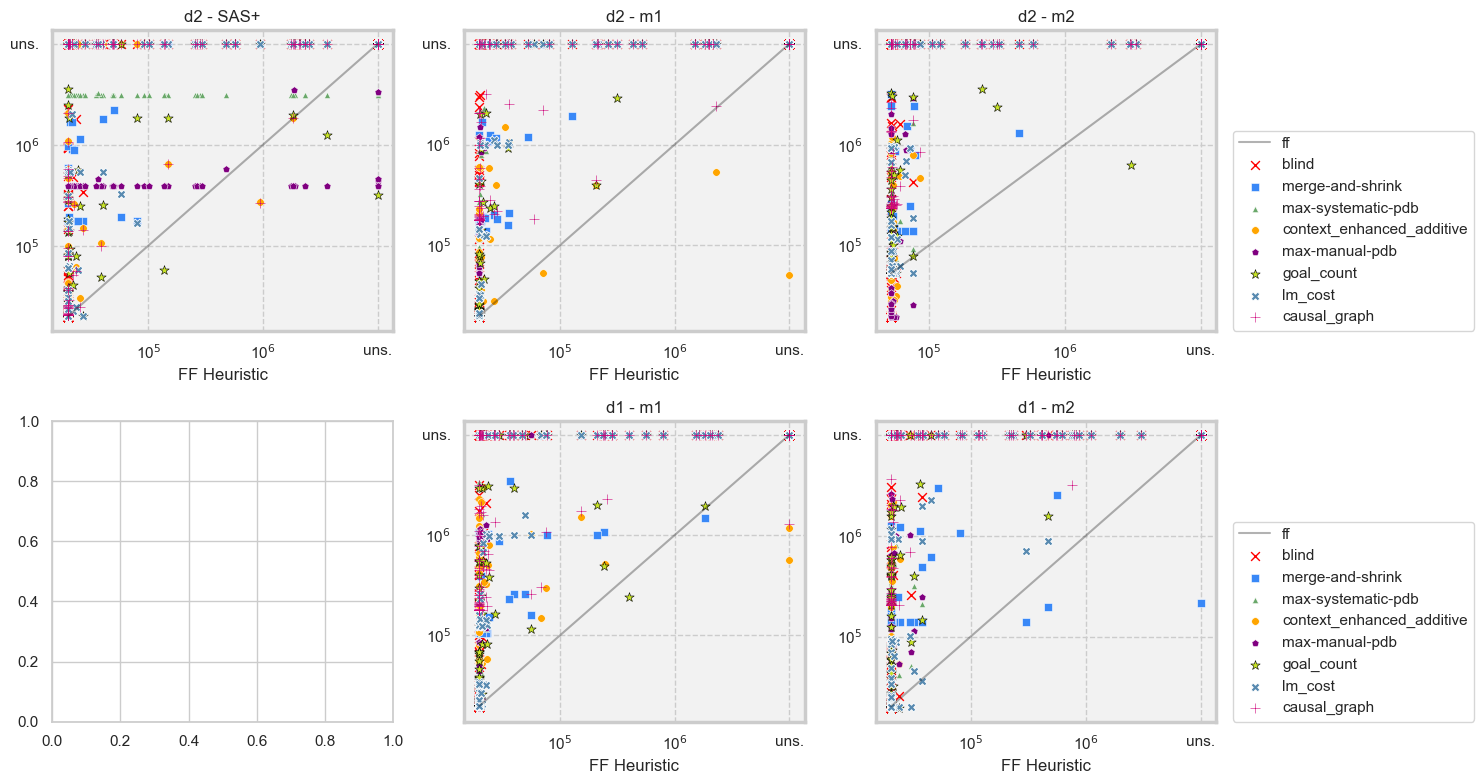}
  \caption{Comparison of memory usage plotted against the FF heuristic}
\end{figure*}

\begin{figure*}[htbp]
  \centering
  \includegraphics[scale=0.5]{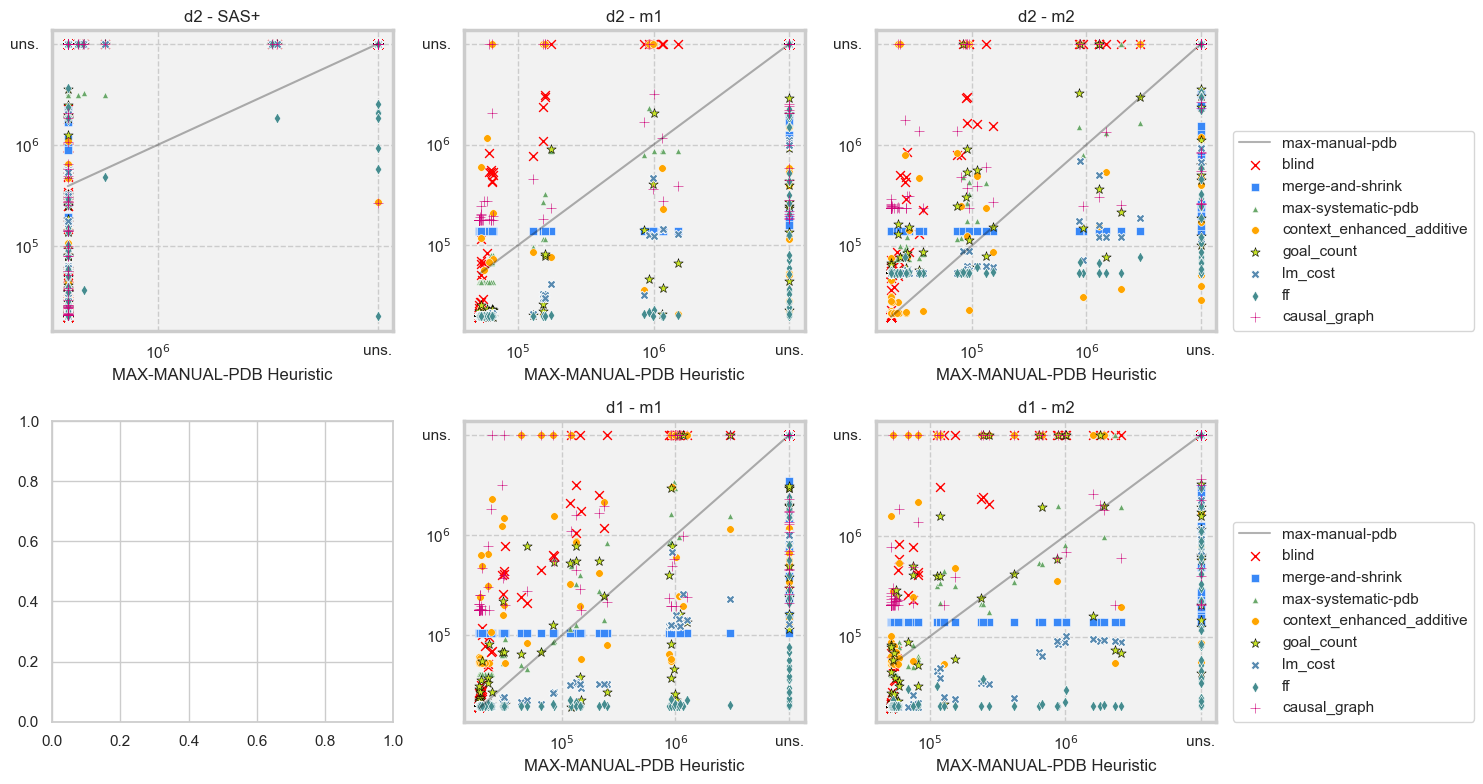}
  \caption{Comparison of memory usage plotted against the Max-Manual-PDB heuristic}
\end{figure*}

\begin{figure*}[htbp]
  \centering
  \includegraphics[scale=0.5]{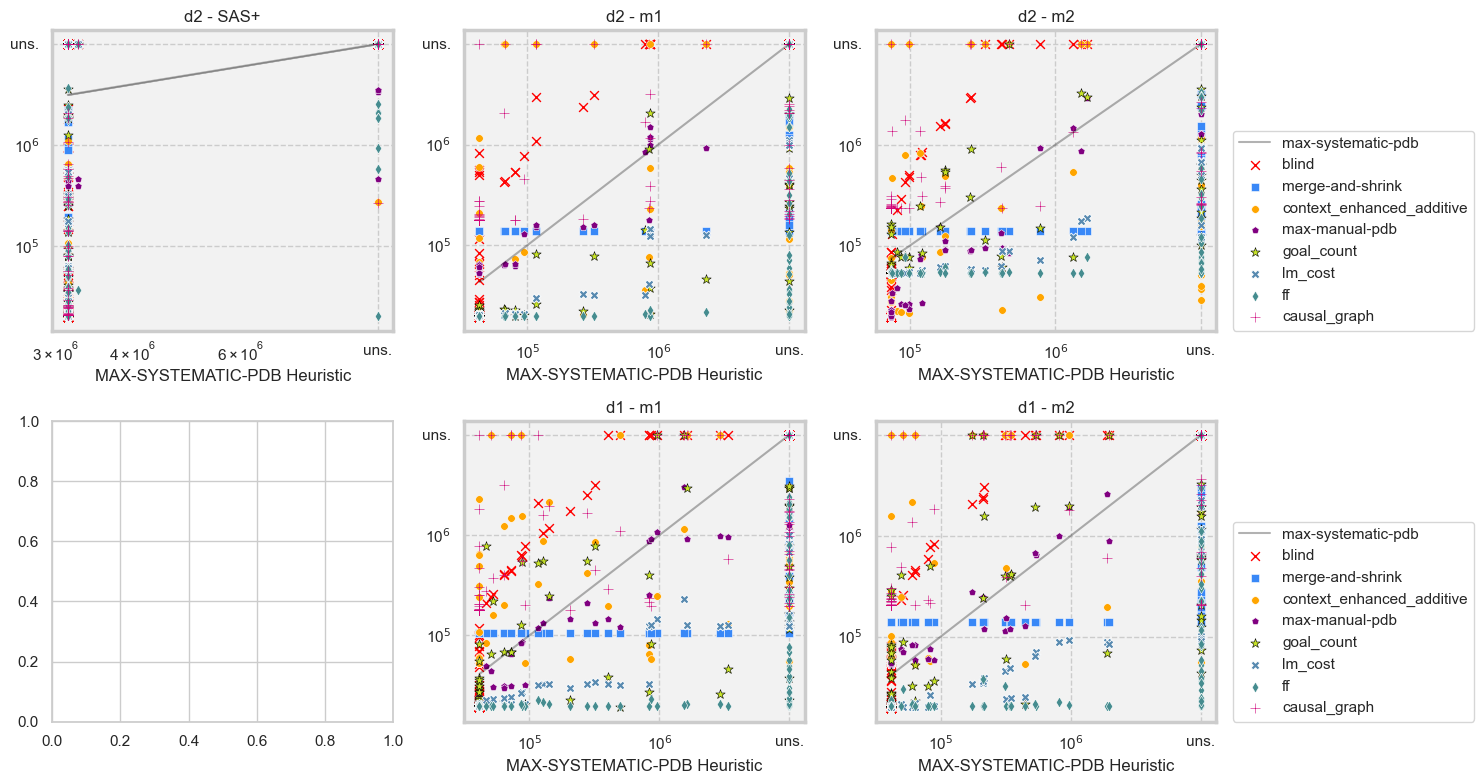}
  \caption{Comparison of memory usage plotted against the Max-Systematic-PDB heuristic}
\end{figure*}

\begin{figure*}[htbp]
  \centering
  \includegraphics[scale=0.5]{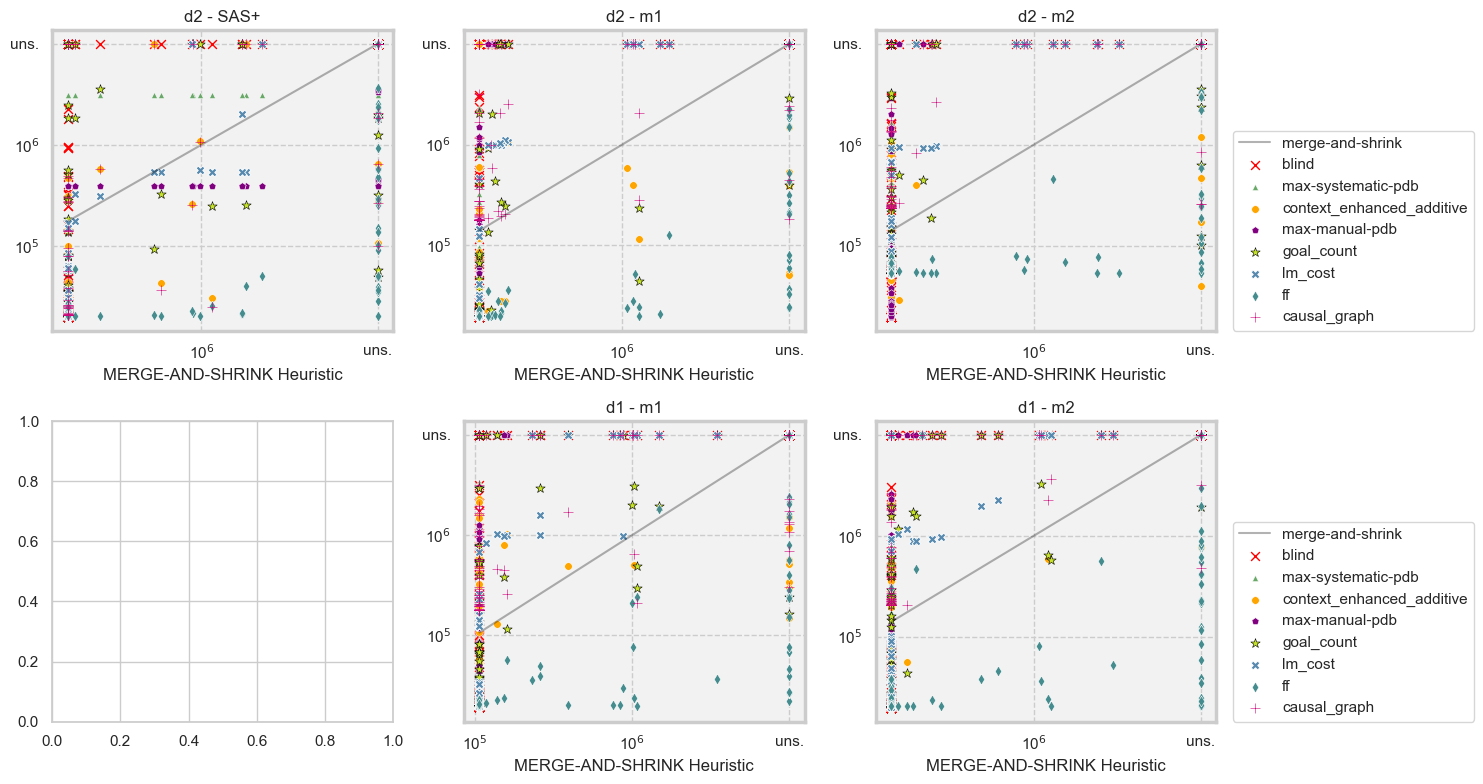}
  \caption{Comparison of memory usage plotted against the Merge-and-Shrink heuristic}
\end{figure*}


\begin{figure*}[htbp]
  \centering
  \includegraphics[scale=0.5]{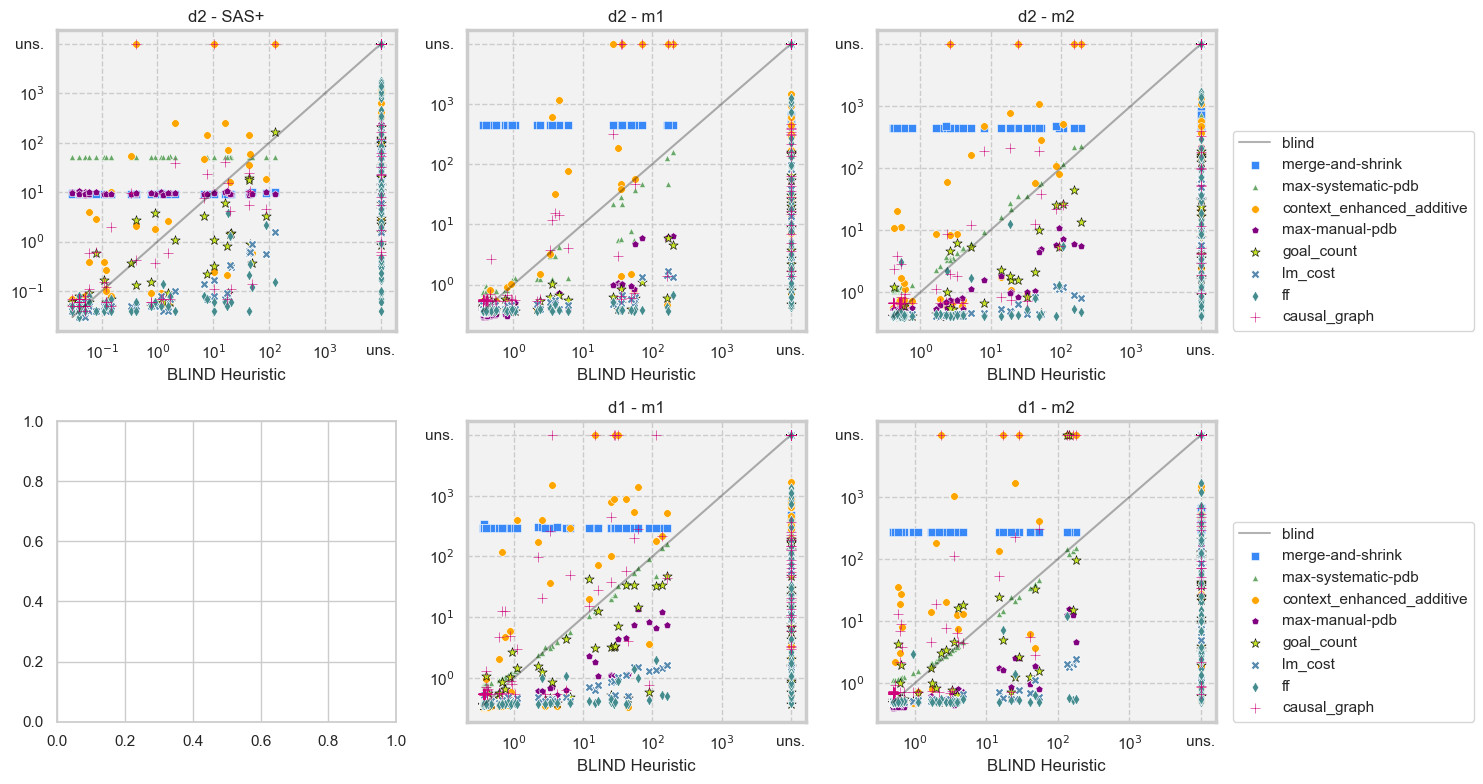}
  \caption{Comparison of runtime plotted against the Blind heuristic}
\end{figure*}

\begin{figure*}[htbp]
  \centering
  \includegraphics[scale=0.5]{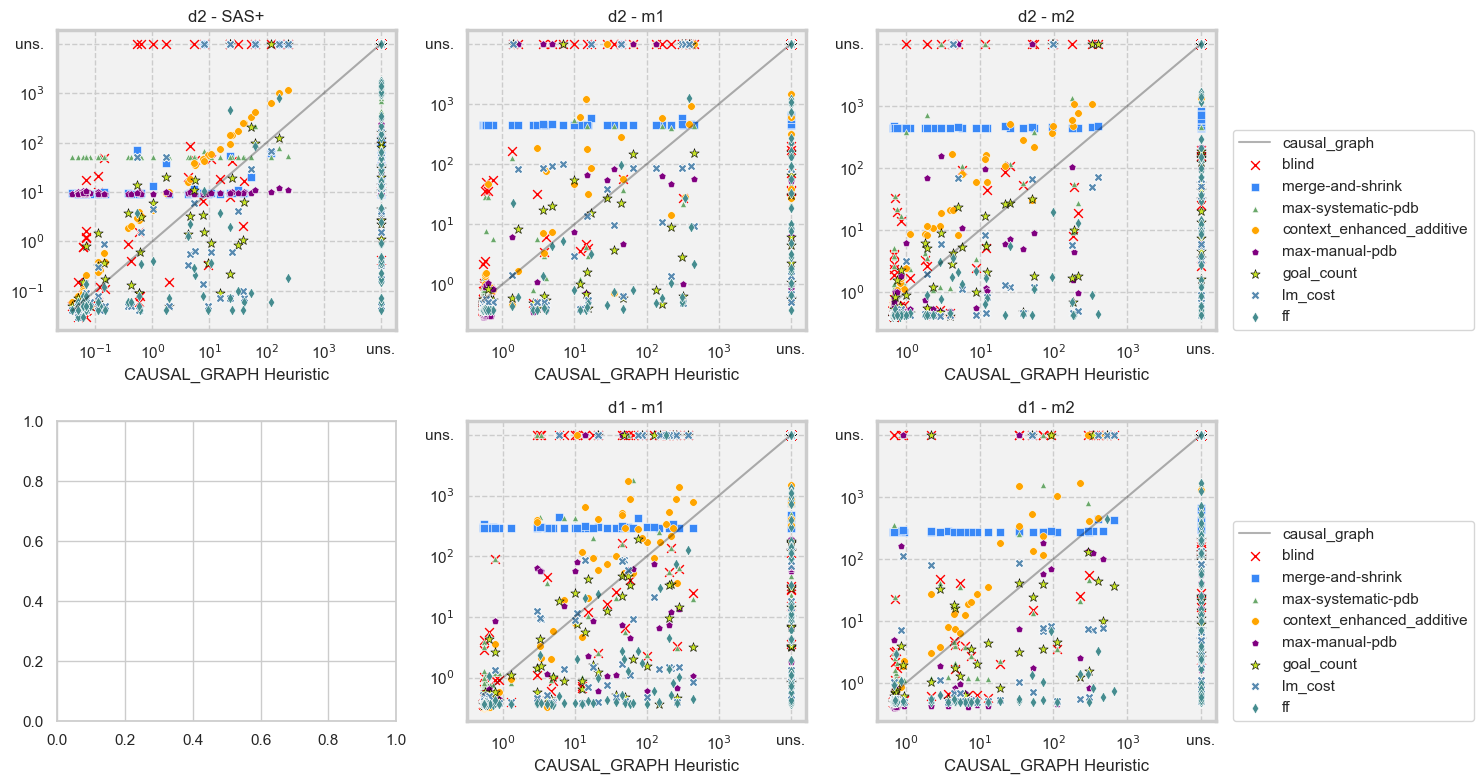}
  \caption{Comparison of runtime plotted against the Causal Graph heuristic}
\end{figure*}

\begin{figure*}[htbp]
  \centering
  \includegraphics[scale=0.5]{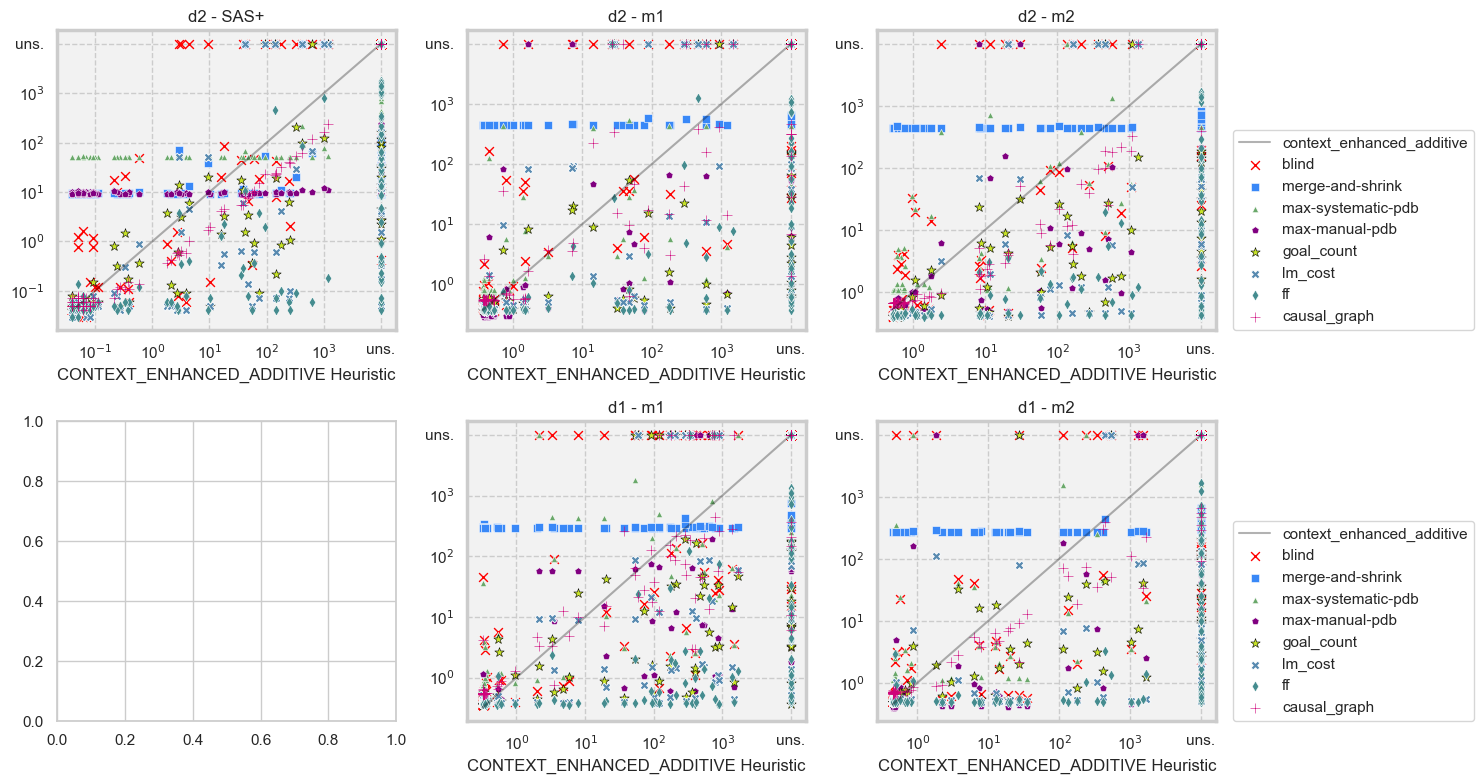}
  \caption{Comparison of runtime plotted against the Context-Enhanced Additive heuristic}
\end{figure*}

\begin{figure*}[htbp]
  \centering
  \includegraphics[scale=0.5]{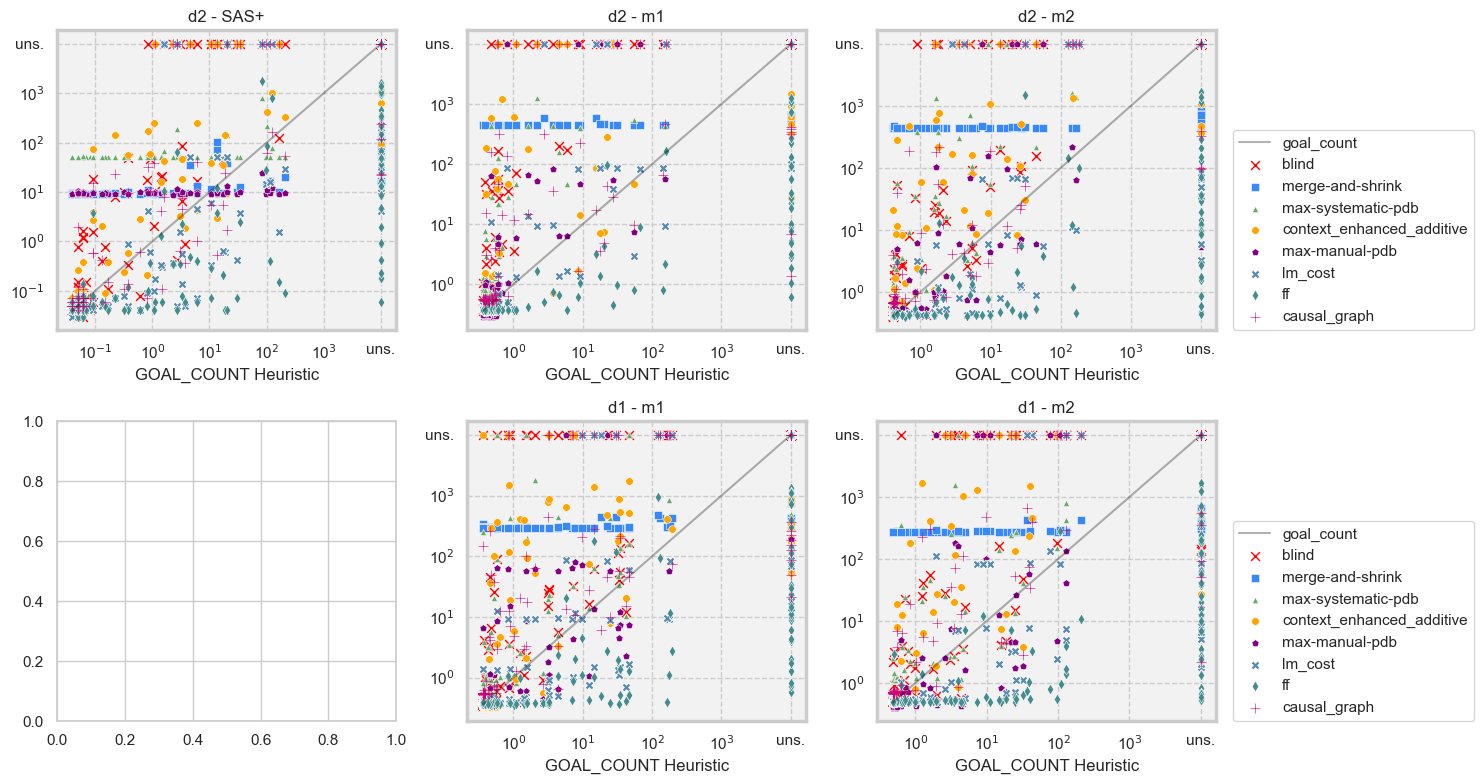}
  \caption{Comparison of runtime plotted against the Goal Count heuristic}
\end{figure*}

\begin{figure*}[htbp]
  \centering
  \includegraphics[scale=0.5]{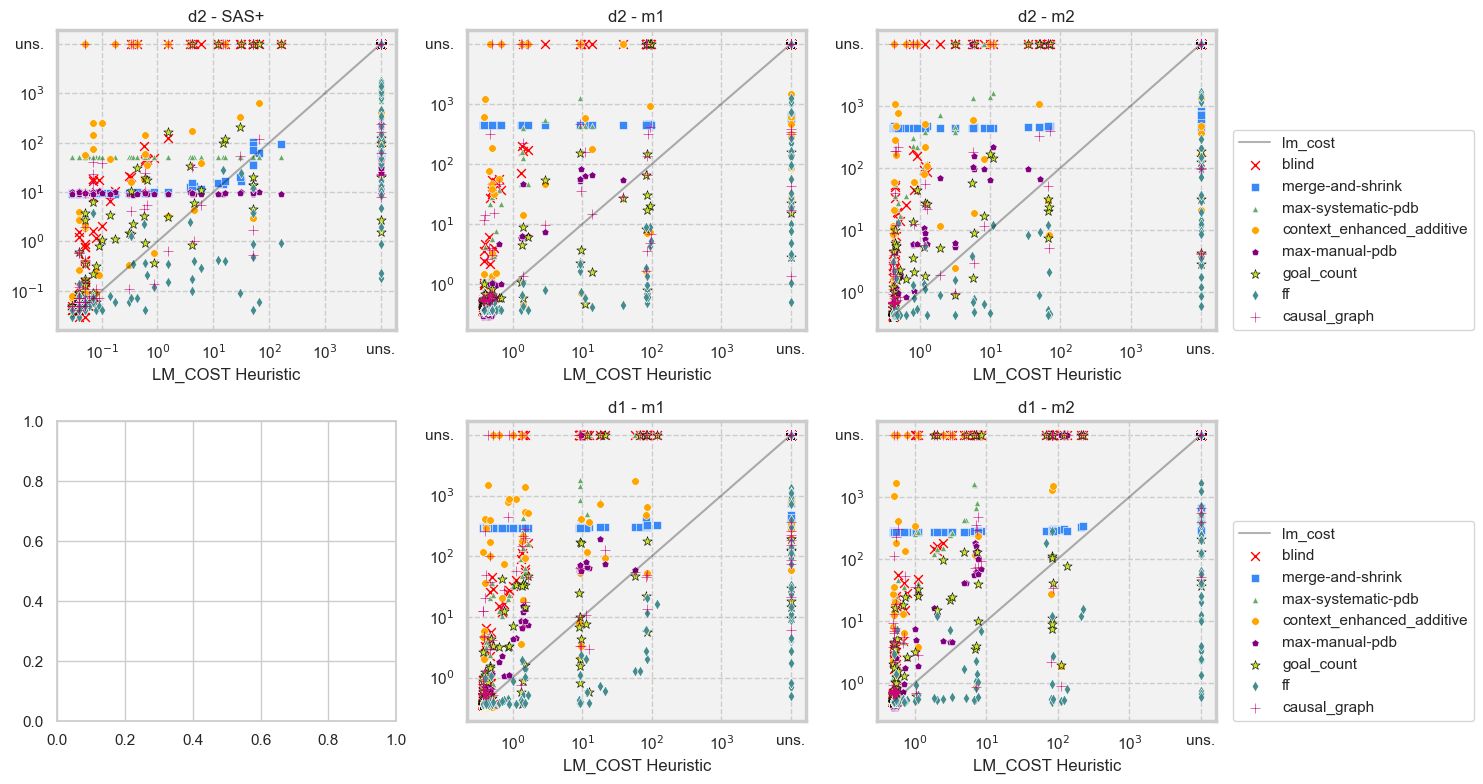}
  \caption{Comparison of runtime plotted against the LM-Cost heuristic}
\end{figure*}

\begin{figure*}[htbp]
  \centering
  \includegraphics[scale=0.5]{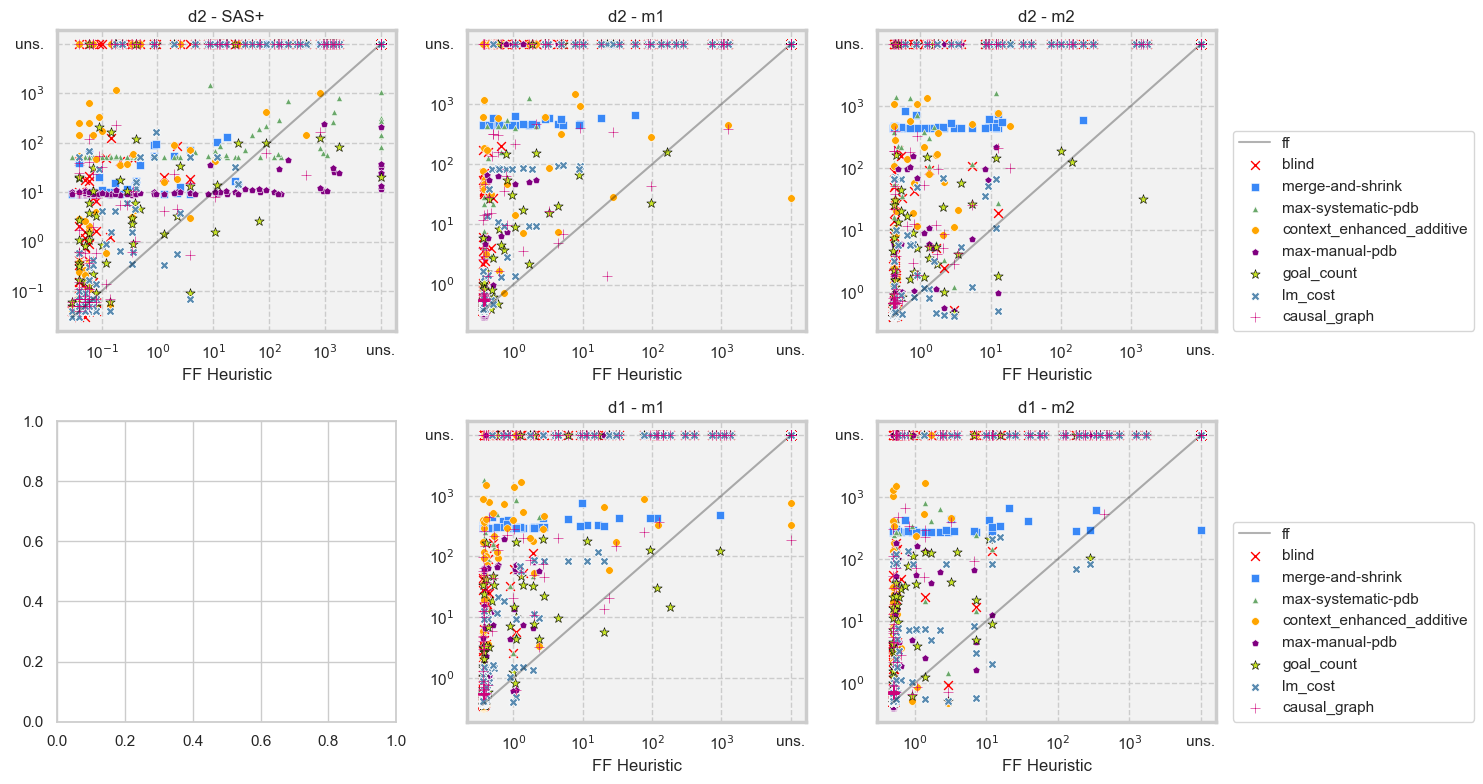}
  \caption{Comparison of runtime plotted against the FF heuristic}
\end{figure*}

\begin{figure*}[htbp]
  \centering
  \includegraphics[scale=0.5]{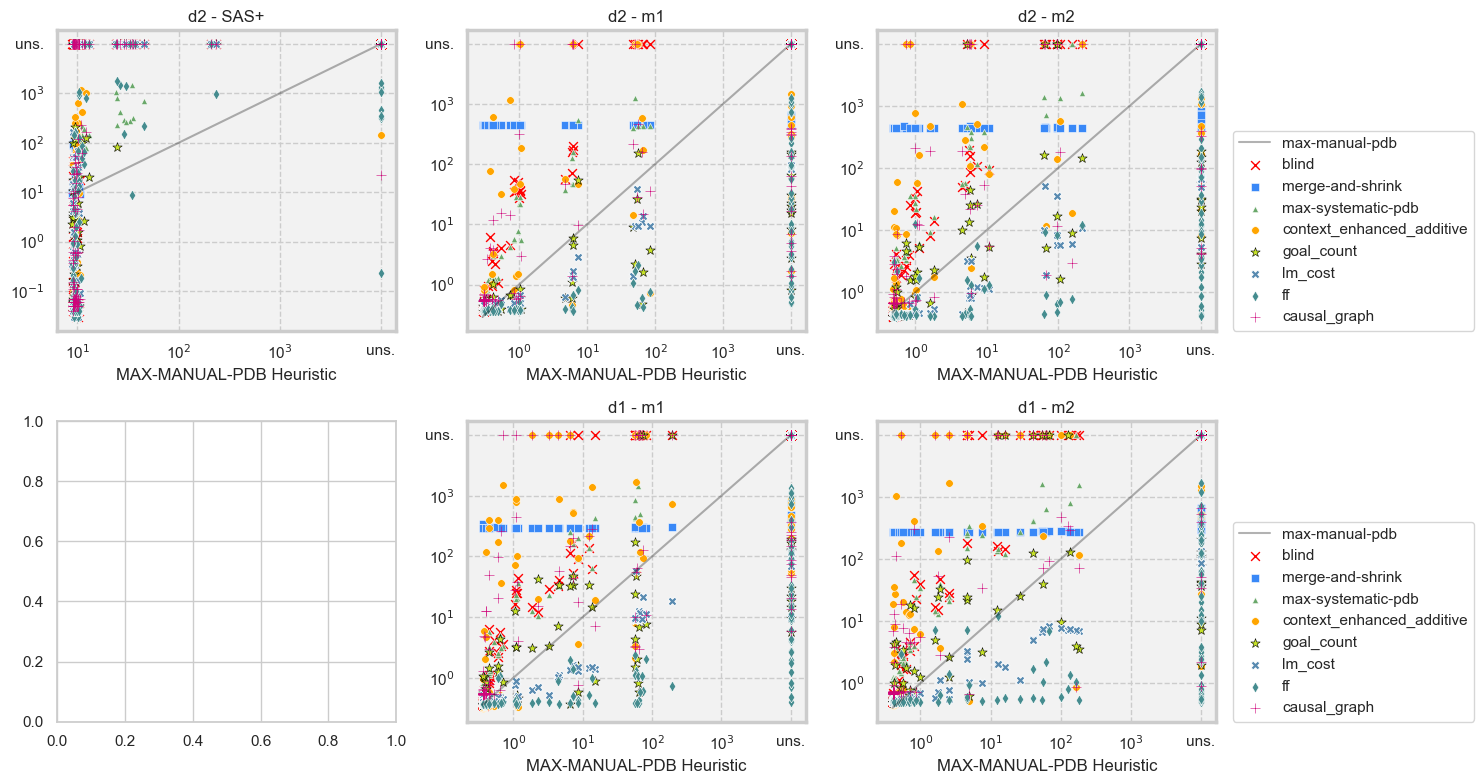}
  \caption{Comparison of runtime plotted against the Max-Manual-PDB heuristic}
\end{figure*}

\begin{figure*}[htbp]
  \centering
  \includegraphics[scale=0.5]{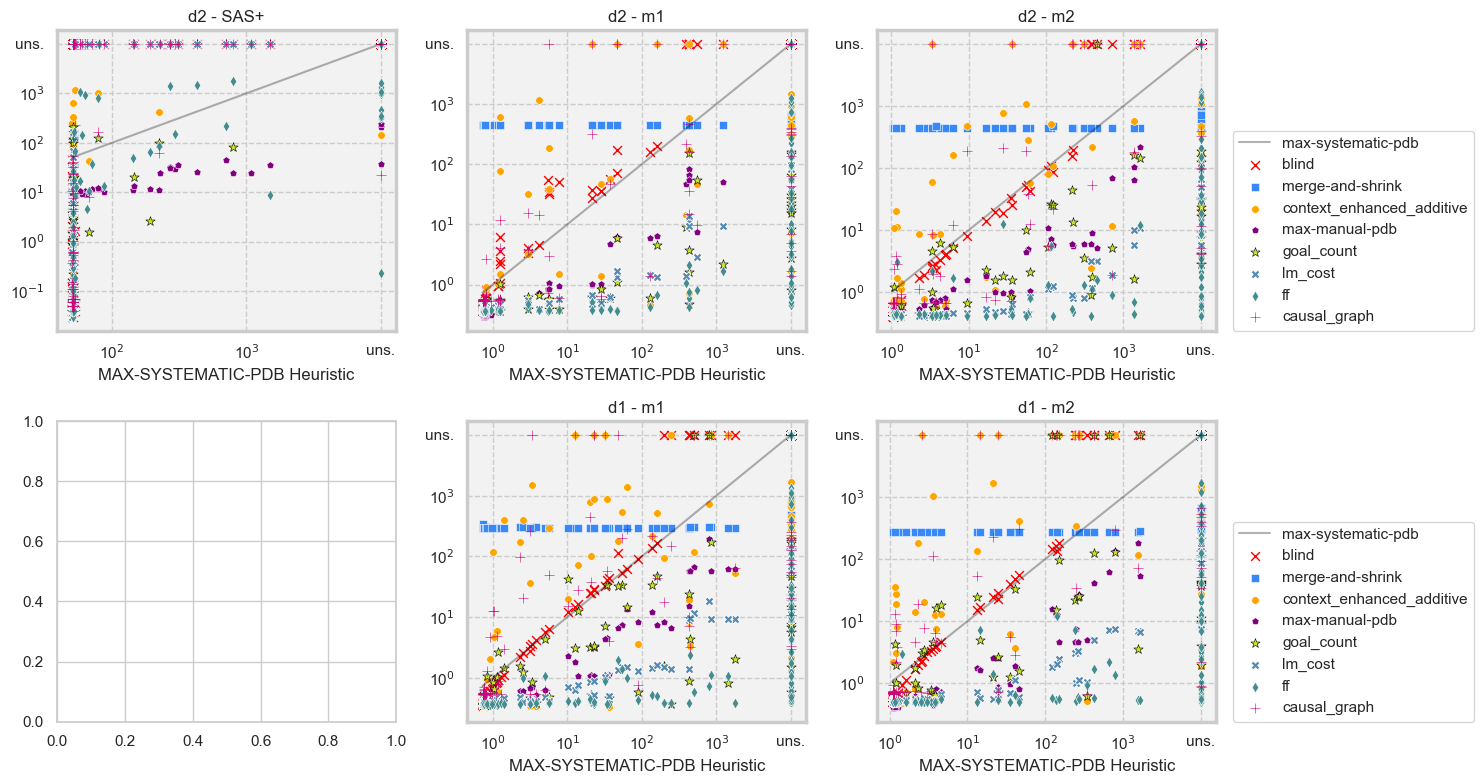}
  \caption{Comparison of runtime plotted against the Max-Systematic-PDB heuristic}
\end{figure*}

\begin{figure*}[htbp]
  \centering
  \includegraphics[scale=0.5]{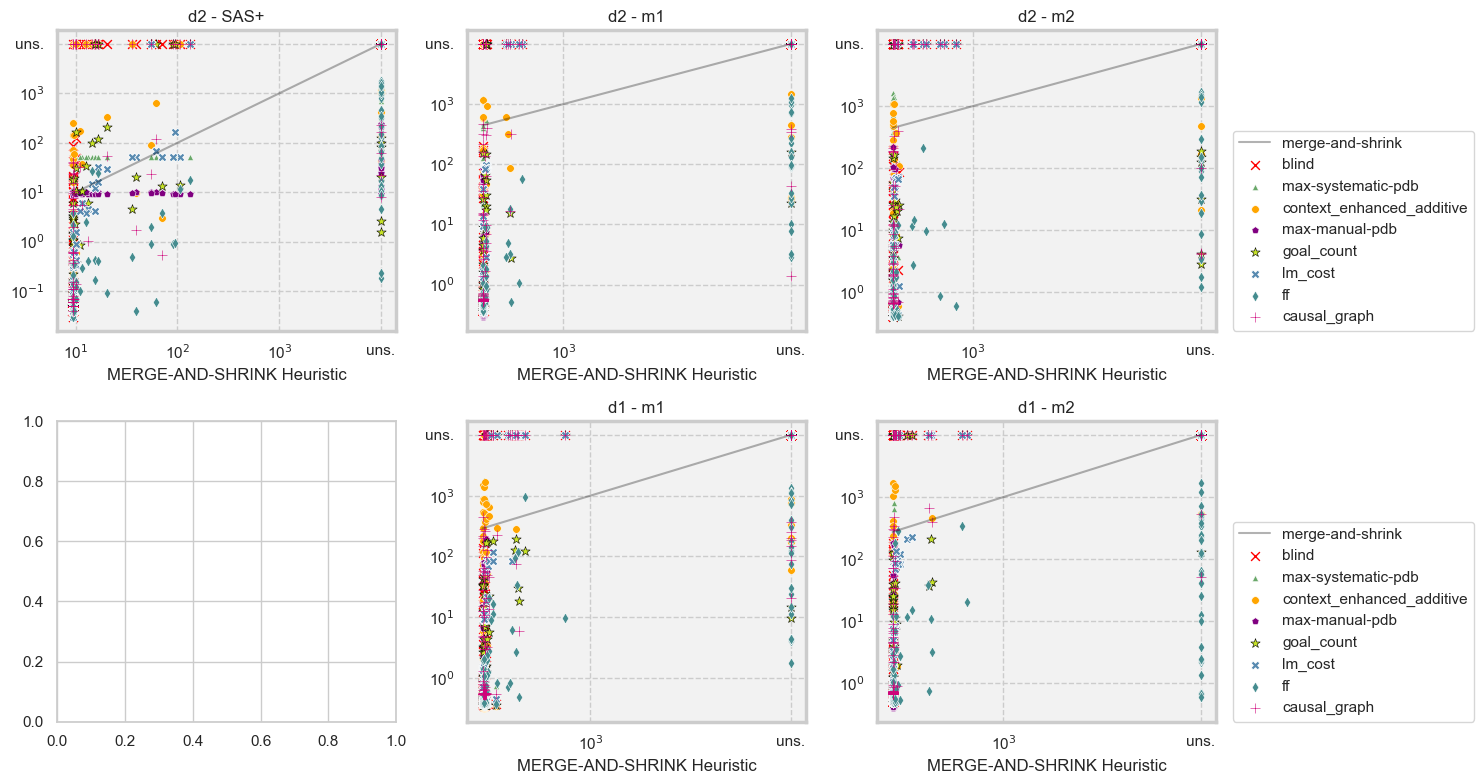}
  \caption{Comparison of runtime plotted against the Merge-and-Shrink heuristic}
\end{figure*}
